\pdfoutput=1  
\documentclass{article}

\usepackage[preprint]{neurips_2026}
\usepackage{amsmath}

\usepackage[utf8]{inputenc} 
\usepackage[T1]{fontenc}    
\usepackage{hyperref}       
\usepackage{url}            
\usepackage{booktabs}       
\usepackage{amsfonts}       
\usepackage{nicefrac}       
\usepackage{microtype}      
\usepackage{xcolor}         
\usepackage{tikz}           
\usetikzlibrary{positioning} 
\usetikzlibrary{arrows.meta} 
\usepackage{longtable}      
\usepackage{array}          
\definecolor{csold}{HTML}{2A78D6}
\definecolor{csnew}{HTML}{EB6834}
\definecolor{csmethod}{HTML}{7A5FC7}
\definecolor{csink}{HTML}{52514E}    

\usepackage{fontawesome5}   
\usepackage[disable]{todonotes}  

\input{notation}  

\newcommand{\cdurlcode}{https://github.com/edwardF8/Circuit-Diff}
\newcommand{\cdurldata}{https://huggingface.co/datasets/edwardF8/Circuit-Diff-paper-data}
\newcommand{\cdurldemo}{https://circuit-diff-paper.vercel.app/}
\newcommand{\cdbadge}[4]{%
  \href{#4}{\tikz[baseline=(bx.base)]{\node[draw=#1,fill=#1!10,rounded corners=3pt,
    inner xsep=7pt,inner ysep=3.2pt,line width=0.6pt](bx)
    {\small\color{#1}#2\hspace{0.4em}\textbf{#3}};}}%
}
\newcommand{\cdbadgerow}{%
  \vspace{-1.15em}%
  {\centering\begingroup\hypersetup{hidelinks}\setlength{\parskip}{0pt}%
  \cdbadge{csold}{\faGithub}{Code}{\cdurlcode}\hspace{0.7em}%
  \cdbadge{csnew}{\faDatabase}{Data}{\cdurldata}\hspace{0.7em}%
  \cdbadge{csmethod}{\faGlobe}{Demo}{\cdurldemo}%
  \endgroup\par}%
  \vspace{0.95em}%
}

\title{Circuit-Diff: Factual Edit-based Intervention Method for Localizing Knowledge in Attribution Graphs}

\author{%
  \begin{tabular}{c@{\hspace{3.5em}}c}
    Edward G.~Friedman\thanks{Correspondence to \texttt{egfriedm@andrew.cmu.edu}.} &
    Xiangchen Song \\[2pt]
    \normalfont School of Computer Science &
    \normalfont Machine Learning Department
  \end{tabular} \\[2pt]
  Carnegie Mellon University \\
  Pittsburgh, PA 15213
}

\begin{document}

\maketitle

\cdbadgerow

\begin{center}
\begin{tikzpicture}[
  font=\scriptsize,
  >={Stealth[length=4pt,width=3pt]},
  mbox/.style 2 args={draw=#1, fill=#1!8, text=#1, rounded corners=2.5pt,
      line width=0.7pt, minimum width=#2, minimum height=8mm, align=center},
  lens/.style={draw=csink, fill=csink!5, text=csink, rounded corners=2.5pt,
      line width=0.7pt, minimum width=20mm, minimum height=30mm, align=center},
  ar/.style={->, line width=0.7pt, draw=csink},
  note/.style={text=csink, font=\scriptsize, align=center},
]
  \node[mbox={csold}{20mm}] (m0) at (0,  1.05) {Original Model\\[-1pt]$\to$ Paris};
  \node[mbox={csnew}{20mm}] (me) at (0, -1.05) {Edited Model\\[-1pt]$\to$ Rome};
  \draw[ar, draw=csmethod] (m0) -- node[left, note, text=csmethod, xshift=1pt]
      {factual\\edit} (me);

  \node[lens] (lens) at (2.70, 0) {Original\\Cross-Layer\\Transcoder\\(CLT)};
  \node[note, text width=50mm] (px) at (2.70, 2.25)
      {\hyphenpenalty=10000\exhyphenpenalty=10000
       ``The Eiffel Tower is located in the city of \underline{\phantom{xxx}}''};
  \draw[ar] (px) -- (lens.north);

  \node[mbox={csold}{27mm}] (ga) at (5.90,  1.05) {Original Attribution\\Graph A};
  \node[mbox={csnew}{27mm}] (gb) at (5.90, -1.05) {Post-Edit Attribution\\Graph B};
  \draw[ar] (m0) -- (m0 -| lens.west);
  \draw[ar] (me) -- (me -| lens.west);
  \draw[ar] (lens.east |- ga) -- (ga);
  \draw[ar] (lens.east |- gb) -- (gb);

  \node[mbox={csmethod}{14mm}] (diff) at (8.65, 0) {$B - A$};
  \draw[ar] (ga.east) -- ++(0.3,0) |- (diff.west);
  \draw[ar] (gb.east) -- ++(0.3,0) |- (diff.west);
  \node[note, below=1.2mm of diff] {circuit-diff};
  \node[mbox={csold}{24mm}] (odis) at (11.05,  1.05) {\textbf{disappeared}\\nodes};
  \node[mbox={csnew}{24mm}] (oapp) at (11.05, -1.05) {\textbf{appeared}\\nodes};
  \draw[ar] (diff.east) -- ++(0.2,0) |- (odis.west);
  \draw[ar] (diff.east) -- ++(0.2,0) |- (oapp.west);
  \node[note, above=1.3mm of odis, text width=32mm, text=csold]
      {features that carried \emph{Paris}, France, French history, \ldots};
  \node[note, below=1.3mm of oapp, text width=32mm, text=csnew]
      {features that carried \emph{Rome}, Italy, Roman mythology, \ldots};
\end{tikzpicture}

\end{center}
\vspace{0.6em}

\begin{abstract}
  Mechanistic interpretability defines \textit{features} as the fundamental units of a
  neural network and \textit{circuits} as the weighted subgraphs that carry out its
  computation. Because individual neurons are polysemantic, Cross-Layer Transcoders
  (CLTs) were introduced as a way to approximate a model's circuits by generating an
  \textit{attribution graph}. The nodes of that graph, however, are unlabeled features:
  reading a graph means pruning it and then working out by hand what each surviving node
  means. To make CLTs easier to use for circuit discovery, we introduce
  \textit{Circuit-Diff}, which intervenes on the model itself with a low-rank factual
  edit and takes the features whose role in the attribution graph changes under that edit
  as related to the edited knowledge. On the edits we examine, the flagged
  nodes are not only detectors of the object token: read off the CLT's released feature
  dashboards, they include features for the history, geography and associations
  surrounding the old and new objects. We formalize the method, measure how reliable a
  frozen CLT remains after a factual edit, test the selected nodes causally by patching
  them on up to 24 CounterFact edits, give a case study, and release an open-source
  implementation built on the \textit{circuit-tracer} package
  \cite{hanna-etal-2025-circuit}, together with two further tools (multi-prompt
  aggregation and rule-based supernode labeling).
\end{abstract}

\section{Introduction}

Early mechanistic interpretability work defined \textit{features} as the fundamental units of a neural network and \textit{circuits} as the weighted subgraphs that represent the computation of the model \cite{olah_zoom_2020}. Yet, studies found that neurons themselves did not correspond to single features, and instead existed in a state of superposition, in which polysemantic neurons would represent many unrelated features (\cite{arora_linear_2018,elhage2022superposition,olah_zoom_2020}). As models grew larger and their internals harder to interpret, Sparse Autoencoders (SAEs) \cite{cunningham_sparse_2023,bricken2023towards} and Cross-Layer Transcoders (CLTs) \cite{persic2025circuittracing} were introduced to approximate such features and circuits. 

However, while SAEs and CLTs proved useful for approximating features and circuits, they are unsupervised and so do not by themselves ascribe semantic meaning to what they find. Early methods relied on manual inspection of where features fired \citep{bricken2023towards,cunningham_sparse_2023}, and automated interpretability methods were introduced to generate descriptions of features \citep{bills_language_2023,paulo_automatically_2024}. These descriptions have been found to be causally unfaithful under interventions in some settings \citep{huang_rigorously_2023}. In the CLT case, the generated \textit{attribution graph} that approximates the circuit can be pruned, but a large number of nodes still need to be labeled. As CLTs and attribution graphs are adopted as a basis for circuit analysis across domains \citep{yang_circuit_2026,kim_craft_2026,mazur_diffract_2026}, there is a need for tools that make this kind of analysis easier and more automated.

We introduce Circuit-Diff, a method that applies a causal intervention to a model via a low-rank knowledge edit using MEMIT \citep{meng_locating_2023,meng_mass-editing_2023} and computes the difference in the attribution graphs \cite{dunefsky_transcoders_2024,persic2025circuittracing} generated by a CLT trained on the unedited model and held fixed on both. The method surfaces feature nodes whose role in the attribution graph changes under the edit, and on the edits we examine those nodes read as the knowledge the edit moved. In this paper, we:
\begin{enumerate}
  \item Formalize the \textit{Circuit-Diff} algorithm, and describe two accompanying
  tools that we release but do not evaluate here: multi-prompt aggregation and, following
  \citet{birardi_automated_2026}, rule-based aggregation of nodes into supernodes, which we do not
  evaluate here.
  \item Measure how reliable a frozen CLT remains after a factual edit, and how many
  feature nodes change role across $N = 500$ CounterFact edits.
  \item Run ablation and steering tests (8--24 CounterFact edits) to check causal
  verifiability, finding that the selected nodes move the target logit substantially more
  than count-matched activation-ranked or random sets; on the inject side they also lead
  a top-influence ranking of the whole graph, while on the suppress side that ranking is
  as good or better.
  \item Show a case study of the nodes the method identifies, read directly from the
  CLT's released feature dashboards.
  \item Release a library and local viewer for applying the method, built on
  circuit-tracer \cite{hanna-etal-2025-circuit}.
\end{enumerate}
\section{Related Work}
\citet{patel_llms_2026} take per-feature descriptions produced by existing auto-interpretability methods (\cite{bills_language_2023}, \cite{paulo_automatically_2024}) and passes them directly to a language model, which groups them into supernodes. However, relying on auto-interpretability descriptions inherits the faithfulness issues \citet{huang_rigorously_2023} report for such methods on SAEs. \citet{arora_adag_2026} introduce ADAG, which characterizes each circuit component by its gradient-based attributions to input tokens and contributions to output logits, clusters functionally similar components into supernodes, and generates and scores natural-language labels for each group with a language model; the pipeline is demonstrated on attribution graphs over MLP neurons but is agnostic to the underlying unit basis. While ADAG replaces the descriptive signal used for grouping, its labels are still produced and scored by language models. 

\citet{birardi_automated_2026} avoid model-generated descriptions altogether: they group features by measuring each feature's activation across a small set of concept-targeted probe prompts and clustering features with similar activation signatures under deterministic rules, validating the resulting supernodes with large-scale interventions. \citet{chen_prune_2026} reduce the number of features prior to interpretation, scoring CLT features by gradient-based attribution to a task metric and retaining the minimal subset that preserves the model's behavior, before running auto-interpretability on the retained set and evaluating both behavioral fidelity and explanation quality.  

Our method shares the goal of \citet{chen_prune_2026} (selecting which features warrant analysis before any interpretation occurs) and shares with \citet{birardi_automated_2026} the commitment to grounding feature analysis in causal effects rather than generated descriptions. In fact, we adopt \citeauthor{birardi_automated_2026}'s procedure to group features by role into supernodes. We differ from both in the form of the intervention and in its position in the pipeline. Where \citet{chen_prune_2026} construct their contrast at the input level, requiring a task-specific clean and corrupted dataset, we construct it at the weight level: a single low-rank factual edit \citep{meng_locating_2023,meng_mass-editing_2023} specifies the counterfactual semantically, with the prompt held fixed. And where \citet{birardi_automated_2026} use interventions to validate a grouping produced upstream, our intervention is the discovery signal itself: the features of interest are those whose role in the attribution graph changes under the edit.

\section{Preliminaries}\label{sec:prelimin}
\paragraph{Factual edits.}
We edit with MEMIT \cite{meng_mass-editing_2023}, the multi-layer successor of
ROME \cite{meng_locating_2023}, on their \textit{CounterFact} dataset. Each
record is a fact $(s, r, o)$ (subject, relation, object) with a prompt $p$
that elicits $o$ from $s$; an edit request $(s, r, o \to o^{*})$ asks the model
to produce a counterfactual object $o^{*}$ of the same relation on $p$ instead.
Each record also supplies \textit{paraphrase} prompts (rephrasings of the same
fact) and \textit{neighborhood} prompts (different subjects sharing $r$ and $o$),
which we use to measure generalization and locality of the edit and, in
\S\ref{sec:meth_Alg}, as the edit's own null.

\paragraph{Attribution graphs.}
A cross-layer transcoder (CLT) $T$ \cite{persic2025circuittracing} replaces
the MLPs of a model $M$ with transcoders \cite{dunefsky_transcoders_2024} to yield a sparse dictionary of features whose decoders write to every later layer. Running $M$ on a prompt $p$ through $T$ yields an
attribution graph $G(M, T; p)$ \cite{dunefsky_transcoders_2024,
persic2025circuittracing}: a DAG whose nodes are token embeddings, active
features, logits, and per-layer \textit{error} nodes (the part of each MLP
output $T$ cannot reconstruct), with edges $w_{n \to n'}$ giving direct linear
attributions between them. We write a feature node as $n = (i, t)$ for feature $i$ at token
position $t$, with activation $a_n$ and \textit{influence} $I_n$: the total normalized
path weight from the target logit ($o$ in the unedited graph, $o^{*}$ in the
edited one) back to $n$, summed over paths of every length on the
row-normalized graph \cite{persic2025circuittracing}. Unlike the standard attribution graph method, which seeds this computation with the probabilities of
all top logits, we seed a single target logit at unit weight, so $I_n$ is the
share of that logit's attribution flowing through $n$ and is comparable across
the two graphs.

\section{Circuit-Diff Method}
\subsection{Circuit-Diff Algorithm}\label{sec:meth_Alg}
Following the notation of \S\ref{sec:prelimin}, given a language model $M_0$ and a
Cross-Layer Transcoder $T_0$, we apply an edit request $(s, r, o \to o^{*})$ to obtain a
model $M_e$ that elicits $o^{*}$ instead of $o$ on prompt $p$. We then compare the
attribution graphs $G(M_0, T_0; p)$ and $G(M_e, T_0; p)$ to find feature nodes and edges
associated with $o$ and $o^{*}$. Throughout we use the running example of
\S\ref{sec:cs_pr}: moving the Eiffel Tower from Paris to Rome, i.e.\
$(\text{Eiffel Tower},\ \text{located in city},\ \text{Paris} \to \text{Rome})$ on the
prompt ``is located in the city of''.

We split our process into five steps:

\textbf{Step 1: Perform the fact edit.} We use MEMIT \citep{meng_mass-editing_2023} to
apply the edit request to $M_0$, producing a model $M_e$ that, when the edit succeeds,
elicits $o^{*}$ (here \texttt{Rome}) on $p$. We prefer multi-layer MEMIT to single-layer ROME
\citep{meng_locating_2023} because factual associations are reported to be distributed
over several layers \citep{geva_dissecting_2023,haviv_understanding_2023,nichani_understanding_2024,persic2025circuittracing};
see Appendix~\ref{apx:memitongemma}. 

\textbf{Step 2: Compute the two attribution graphs and their difference.} Given $M_0$ and
$M_e$, we compute $A = G(M_0, T_0; p)$ and $B = G(M_e, T_0; p)$, using the same
CLT on both so that node identities are comparable, an assumption
\S\ref{sec:cltReliablityafteredit} tests and \S\ref{sec:limitations} qualifies. For every node
$n = (i,t)$, i.e.\ $T_0$'s feature $i$ at token position $t$, we compute the change in the
node's influence on the target logit ($o$'s logit node in $A$, $o^{*}$'s in $B$) and
$\delta_n$, which tracks the normalized movement of its activation. For each edge we
compute the change in edge weight and an influence-weighted distribution over $n$'s
outgoing edges. All quantities are defined in Table~\ref{tab:node-diffs}.

\textbf{Step 3: Track appearing and disappearing nodes.} We use $\delta_n$ from
Table~\ref{tab:node-diffs} to characterize each feature's behavior under the diff. A node
$n$ \textbf{increased} if $\delta_n > 0$ and \textbf{decreased} if $\delta_n < 0$. The
extremes are the candidates of interest: a node \textbf{disappeared} if $\delta_n = -1$,
making it a candidate for a node associated with predicting $o$, and \textbf{appeared} if
$\delta_n = 1$, making it a candidate for a node associated with predicting $o^{*}$.

\textbf{Step 4: Apply a null calibration.} As \S\ref{sec:exper_featurefound} shows, a
large fraction of nodes fall into the appeared and disappeared sets, so a further filter
is useful when a more constrained set is wanted. We compute a floor
$\epsilon = Q_p\big(\big\{\,\lvert a_n^{(B)} - a_n^{(A)}\rvert \;:\; n \in {\mathrm{NBR}}\,\big\}\big)$
from the neighborhood prompts (which elicit unedited knowledge and so serve as the
edit's own null), where $p$ is the chosen percentile, and keep on each side only
the nodes whose movement exceeds $\epsilon$. Every ablation in \S\ref{sec:exper_verif}
is reported both with and without this floor.


\textbf{Step 5: Rank the nodes on each side.} Since we hypothesize that increased and
decreased nodes correspond to $o^{*}$ and $o$ respectively, we split the nodes into two
rankings, one per behavior, and score each with a heuristic meant to surface the features
most associated with the change. We rank each side with $\phi^{(a)}_n = |\delta_n|^{\alpha}\,|\Delta I_n|$, the weighted
logit change ($\alpha$ per Appendix~\ref{apx:defaults}), and measure it against two
method-free reference baselines over the same nodes: activation,
$\phi_{\mathrm{act},n} = |a_n^{\mathrm{ref}}|$ (the activation $\lambda$ scales), and
influence, $\phi_{\mathrm{inf},n} = \max(|I_n^{(A)}|,|I_n^{(B)}|)$. All are unsigned
ranking strengths; direction comes from $\operatorname{sign}(\delta_n)$, which splits the
nodes into appeared and disappeared before ranking. Five further scores we tried are
defined in Table~\ref{tab:scores_extra}; \S\ref{sec:exper_verif} finds four of them
indistinguishable from $\phi^{(a)}$ and one, $\phi^{(e)}$, no better than random.

\begin{table}[t]
  \caption{Per-node difference quantities extracted from the attribution-graph
  pair $A=G(M_0, T_0; p)$ and $B=G(M_e, T_0; p)$, for each node $n=(i,t)$
  (feature $i$ at token position $t$) and edge $n \to n'$.}
  \label{tab:node-diffs}
  \centering
  \small
  \begin{tabular}{@{}llp{5.6cm}@{}}
    \toprule
    Quantity & Definition & Measures \\
    \midrule
    $\Delta a_n$ & $a_n^{(B)} - a_n^{(A)}$ &
      Change in activation of feature $i$ at token position $t$. \\
    $\Delta I_n$ & $I_n^{(B)} - I_n^{(A)}$ &
      Change in node $n$'s influence on the target logit
      ($o^*$'s logit node in $B$, $o$'s in $A$). \\
    $\Delta w_{n\to n'}$ & $w_{n\to n'}^{(B)} - w_{n\to n'}^{(A)}$ &
      Change in edge weight from node $n$ to node $n'$. \\
    $\pi_n(n')$ &
      $\propto |w_{n\to n'}| \cdot \max\bigl(|I_{n'}^{(A)}|, |I_{n'}^{(B)}|\bigr)$ &
      Importance-weighted distribution of $n$'s outgoing edge mass, normalized
      over its targets; \\
      $\delta_n$ &
      $\Delta a_n \,/\, |\max\bigl(a_n^{(B)},\, a_n^{(A)},\, \varepsilon_n\bigr)|$ &
      Normalized activation movement, in $[-1,1]$; the guard
      $\varepsilon_n = \tau_a \cdot a_i^{\max}$ (feature-level maximum of $n$'s
      feature $i$) keeps near-silent nodes from saturating the ratio. \\
    \bottomrule
  \end{tabular}
\end{table}

\subsection{Circuit-Diff Platform and Tools} \label{sec:meth_Platform}
Along with this paper, we release the Circuit-Diff Platform, which builds on the
circuit-tracer package \cite{hanna-etal-2025-circuit} to compute the diff,
inspect it visually and export the tables and figures used here
(Appendix~\ref{apx:circuitdiffplatform}). Two further tools ship with it and are not
evaluated here: \emph{multi-prompt aggregation}, which pools a feature's score over
rephrasings of the same fact (Appendix~\ref{apx:mpd}), and \emph{supernode labeling},
which sorts the returned features into roles by their activations on a fixed probe set,
with no language-model-written descriptions, following
\citet{birardi_automated_2026} (Appendix~\ref{apx:autointerp}).

\section{Experiments}
All experiments use Gemma-2-2B \citep{team_gemma_2024} and the released 426k-feature CLT
of \citet{clt_gemma2_2b_426k}. We refer to four model--CLT combinations on a given fact:
$A$ and $B$ from \S\ref{sec:meth_Alg}, plus the adapted-lens graphs
$C = G(M_e, T_f; p)$ and $D = G(M_0, T_f; p)$, where $T_f$ is the CLT fine-tuned on
$M_e$ (Appendix~\ref{apx:clt_finetuneing}). Each pair isolates one factor: $B - A$ is the
circuit-diff; $C$ versus $B$ isolates lens adaptation with the model fixed at $M_e$; and
$D$ is the control corner: the fine-tuned lens on the \emph{unedited} model, i.e.\ what
fine-tuning alone does absent any edit. Compute is reported in
Appendix~\ref{apx:compute}.

\subsection{Cross-Layer Transcoders Reliability After Factual Edit} \label{sec:cltReliablityafteredit}
\begin{figure}[h]
    \centering
    \includegraphics[width=\textwidth]{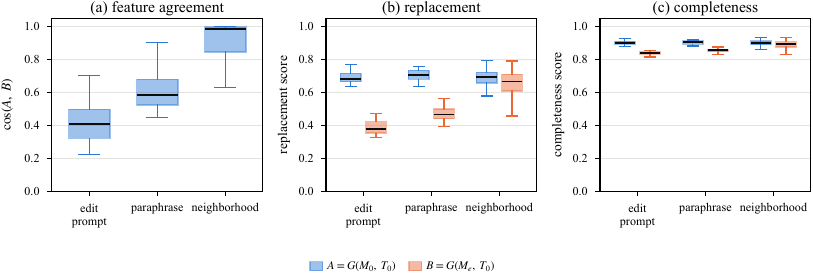}
    \caption{A single MEMIT edit damages the frozen lens exactly where it lands; while the replacement score drops, the less strict completeness score isn't as low in the edited case.  Per-prompt agreement and graph quality for $A = G(M_0, T_0)$ versus $B = G(M_e, T_0)$. A single fact is edited per model ($k = 1$) over $n = 20$ trials; within each trial we
split by prompt bucket, where edit and paraphrase both elicit the edited fact and
neighborhood elicits unedited facts.}
    \label{fig:single_edit_A_vs_B}
\end{figure}

To test whether $T_0$ survives the edit we compare the completeness score, the
replacement score and per-layer cosine activation similarity between $A$ and $B$ over
the edit, paraphrase and neighborhood prompts of \S\ref{sec:prelimin}
(Figure~\ref{fig:single_edit_A_vs_B}). The damage to the lens is localized to the edit. The replacement score of the attribution
graph falls sharply on edited prompts (median drop ${\sim}0.3$) while the less strict
completeness score falls much less (median drop ${\sim}0.06$); on neighborhood prompts
neither moves appreciably. Because the drop appears on the edit and paraphrase buckets
but not on the neighborhood bucket, we attribute it to the knowledge edit, and read the
gap between the two scores as the CLT's error nodes absorbing what the lens can no longer
model. This remains a core limitation (\S\ref{sec:limitations}). We deliberately do not
fine-tune the base CLT, since doing so would break the correspondence between its
features and the released feature dashboards we read them from.

We also fine-tuned the CLT on $M_e$'s activations over general wiki text ($T_f$); the fine-tune early-stops at parity and does not recover the replacement score, so the frozen lens $T_0$ is used throughout (Appendix~\ref{apx:clt_finetuneing}).

\subsection{Circuit-Diff Nodes that Appeared/Disappeared}\label{sec:exper_featurefound}
\begin{figure}[t]
    \centering
    \includegraphics[width=\textwidth]{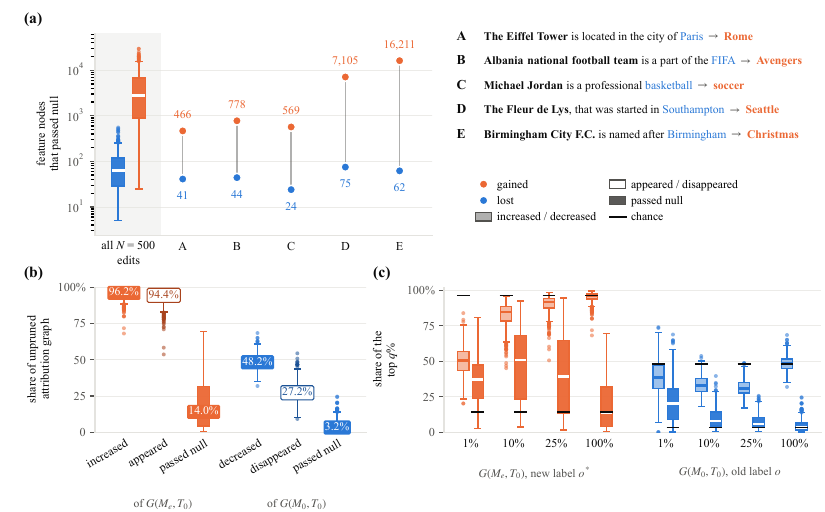}
    \caption{%
        \textbf{Feature nodes changed by a knowledge edit, under three readings.}
        $N = 500$ CounterFact edits, $G(M_0, T_0)$ vs.\ $G(M_e, T_0)$ on the
        edited prompt. Direction (increased/decreased), presence
        (appeared/disappeared), and \emph{passed null}: moved by more than the
        edit's own control-prompt floor, the $99$th percentile of raw
        $|\Delta a_n|$ (\S\ref{sec:meth_Alg}, Step~4).
        \textbf{(a)} Passed-null nodes per edit, each against its own floor, so
        counts are not comparable across edits; A is a case study, not
        among the $N$.
        \textbf{(b)} Each side as a share of its own graph.
        \textbf{(c)} Share of the top-$q$\% by influence on the target logit,
        direction vs.\ passed null. The two sides use different rankings
        (gained: $G(M_e, T_0)$ against $o^{*}$; lost: $G(M_0, T_0)$ against
        $o$) and are not comparable; black rules are each series' chance level.%
    }
    \label{fig:setdiff_combined_null}
\end{figure}

How much of a graph does one edit move? Figure~\ref{fig:setdiff_combined_null} answers
this over $N = 500$ CounterFact edits under three readings of ``changed'': direction,
presence, and \emph{passed null} (Step~4, \S\ref{sec:meth_Alg}). The first two are
close to uninformative on their own: in panel (b) the gained side covers $96.2\%$ of the
graph by direction and $94.4\%$ by presence, and the lost side $48.2\%$ and $27.2\%$.
Applying the floor collapses these to $14.0\%$ and $3.2\%$ respectively. Panel~(c) shows
the same effect against the influence ranking: unfloored, the increased share rises with
$q$ but stays below its chance level throughout, whereas floored it sits $2$--$3\times$
above its $14.0\%$ chance rule at every $q < 100\%$.

\subsection{Verification of Circuit-Diff Nodes} \label{sec:exper_verif}
\paragraph{Patching protocol.}
To verify the selected nodes causally we patch them in the unedited model. Since each
node also feeds later layers, we use \citeauthor{persic2025circuittracing}'s constrained
patching: given a node $n = (i,t)$, we clamp its value to $\lambda \cdot a_n$ and add it
to the original MLP output for a window of layers ahead of it, sweeping the window
endpoint. Each patch is applied to $G(M_0, T_0; p)$: appeared nodes are scored on whether
they reproduce the effect of the edit (raising the $o^{*}$ logit), disappeared nodes on
whether they suppress the old answer (lowering the $o$ logit). We score a patch by $R^{z}$, its effect as a fraction of the edit's own on the
centered-logit basis, so $R^{z} = 1$ reproduces the edit exactly ($\varepsilon_z = 0.5$
guards the denominator):

\begin{equation}
    R^{z}_{t} = \frac{z_t^{\mathrm{patch}} - z_t^{(M_0)}}
           {\max\!\big(\big|z_t^{(M_e)} - z_t^{(M_0)}\big|,\ \varepsilon_z\big)},
    \qquad z_t = \ell_t - \frac{1}{|V|}\sum_{v \in V} \ell_v .
    \label{eq:rz}
\end{equation}

\paragraph{Selection and floor policy.} All arms rank with $\phi^{(a)}$
(\S\ref{sec:meth_Alg}, Step~5); at the depths we patch (top $10\%$ at most) every
selected node is an appeared/disappeared node, not merely an increased/decreased
one. We report every ablation under both calibrations: unfloored, and restricted to the nodes that pass the $p=99$ null floor of Step~4 (median survival $13\%$ of the increased side and $8\%$ of the decreased side); under the floor, $q$ is a share of the survivors. Every arm is compared against two method-free baselines that rank \emph{every} node of
the graph by influence or activation on that side's own target
($\phi^{\mathrm{graph}}_{\mathrm{inf}}$, $\phi^{\mathrm{graph}}_{\mathrm{act}}$) and
against a random draw, with counts matched at every rung, so a gap between arms is which
nodes were selected and never how many.

\paragraph{Single nodes.} As \citet{persic2025circuittracing} report, one
feature node rarely moves the logit on its own, because a concept is split
across several features; the per-node sweep is in
Appendix~\ref{apx:ablation_extra} (Figure~\ref{fig:ablate_single}). Per node, the whole-graph influence ranking is ahead of $\phi^{(a)}$ on the suppress side
at every rung under both calibrations and on the unfloored inject side; only on the
floored inject side at $q \ge 5\%$ does $\phi^{(a)}$ lead. The activation baseline falls
to random once it cannot inherit the diff's partition. These are pooled medians over 8
edits, and no per-edit sign test is available at the single-node level, so they should be
read as descriptive.

\begin{figure}[h]
    \centering
    \includegraphics[width=\textwidth]{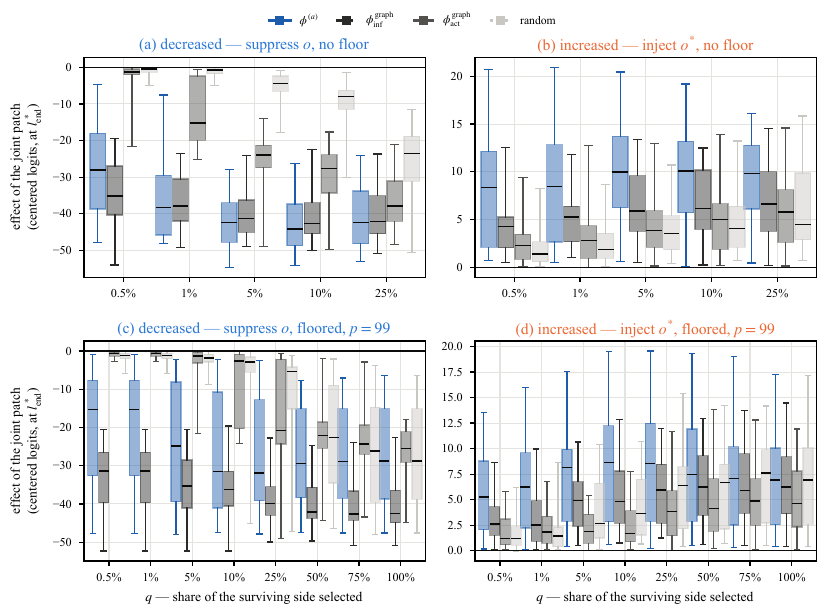}
    \caption{%
        \textbf{Group ablation: the whole selected set patched in one pass,
        under both calibrations.}
        Every member of the top-$q$\% set is clamped simultaneously; arms,
        count matching and the $l^{*}_{\mathrm{end}}$ criterion as in
        Fig.~\ref{fig:ablate_single}, each set read at its own best
        $(\lambda, l^{*}_{\mathrm{end}})$. Top row: no floor. Bottom row:
        $p=99$ survivors; its ladder runs to $q = 100\%$, where every arm
        patches the same survivor set and the arms coincide. Boxes are 24
        CounterFact edits, box = IQR, whiskers = full range.%
    }
    \label{fig:ablate_group_k}
\end{figure}

\paragraph{Groups.} To see the effect of the set the method returns, we patch
the whole top-$q\%$ set in one pass (Figure~\ref{fig:ablate_group_k}). At $q = 5\%$ the selected set moves the target logit further than a count-matched
whole-graph influence ranking on the inject side ($9.98$ versus $5.90$ centered logits
unfloored, $8.16$ versus $4.94$ floored); on the suppress side the two are comparable
unfloored ($-42.4$ versus $-41.3$) and the baseline is ahead under the floor ($-24.8$
versus $-35.3$). Grouping by \textit{supernode}, i.e.\
by shared meaning or role, would likely be a sharper unit of intervention than the top
$q\%$ of a side taken as a whole; we leave that comparison to future work.

\paragraph{Which score.} Widening the group ablation to all six scores (Table~\ref{tab:scores_extra} and Figure~\ref{fig:ablate_scores_k}, Appendix~\ref{apx:ablation_extra}) changes nothing: $\phi^{(a)}$--$\phi^{(d)}$ are indistinguishable because each multiplies influence by $|\delta_n|$, which is saturated at $1$ for nearly every node on both sides, and $\phi^{(e)}$, the only edge-only score, tracks random. Against the whole-graph influence baseline, per-edit sign tests at the $0.156$
replicate-noise threshold favor $\phi^{(a)}$ on the unfloored inject side on 9--11 of the
12 edits at every $q$; on the suppress side the result is mixed, and at $q \le 1\%$ the
baseline is ahead.


\section{Case Studies}\label{sec:cs}
\begin{figure}[t]
  \centering
  \begin{minipage}[t]{0.495\textwidth}\centering
    \includegraphics[width=\linewidth]{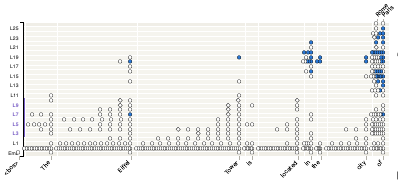}\\[-1pt]
    {\small (a) $A$: disappeared nodes in \textcolor{csold}{blue}}
  \end{minipage}\hfill
  \begin{minipage}[t]{0.495\textwidth}\centering
    \includegraphics[width=\linewidth]{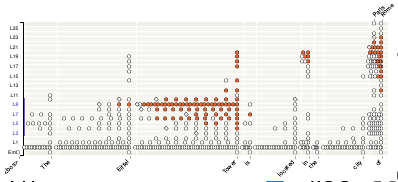}\\[-1pt]
    {\small (b) $B$: appeared nodes in \textcolor{csnew}{orange}}
  \end{minipage}\\[5pt]
  {\scriptsize\setlength{\tabcolsep}{3pt}
\definecolor{csshared}{HTML}{8A8985}
\setlength{\fboxsep}{0.6pt}%
\begin{tabular}{@{}l@{\hspace{5pt}}>{\raggedright\arraybackslash}p{6.75cm}>{\raggedright\arraybackslash}p{3.1cm}@{}}
\multicolumn{3}{@{}l}{\textbf{Top readouts onto ``Paris'' in $A$ (base model)}}\\[1pt]
\toprule
{\color{csold}\rule[-2.5pt]{2.2pt}{9pt}}\,\textbf{\#1\ L18/f7269}\,{\color{black!60}$+1.61$}\,{\color{csold}$\phi$\,1} & \ldots{}world (British Museum at London\colorbox{csold!8}{\strut ,} \colorbox{csold!14}{\strut Grand} \colorbox{csold!37}{\strut Palais} \colorbox{csold!70}{\strut \textbf{at}} \colorbox{csold!8}{\strut Paris}\colorbox{csold!8}{\strut ,}\ldots{} & {\tiny le, É, Ré, dé, les}\\
{\color{csold}\rule[-2.5pt]{2.2pt}{9pt}}\,\textbf{\#2\ L20/f9908}\,{\color{black!60}$+1.23$}\,{\color{csold}$\phi$\,2} & \ldots{}sculpture can be seen at The Louvre Museum \colorbox{csold!70}{\strut \textbf{in}} \colorbox{csold!8}{\strut Paris}\colorbox{csold!34}{\strut ,}\ldots{} & {\tiny normaux, TagMode, financières, parsedMessage, sauvages}\\
{\color{csshared}\rule[-2.5pt]{2.2pt}{9pt}}\,\textbf{\#3\ L19/f9768}\,{\color{black!60}$-1.06$} & University of Tokyo\colorbox{csshared!8}{\strut \P{}}7-3-1 H\colorbox{csshared!9}{\strut ongo}\colorbox{csshared!26}{\strut ,} Bunkyo-\colorbox{csshared!46}{\strut ku}\colorbox{csshared!70}{\strut \textbf{,}} Tokyo 1\ldots{} & {\tiny úgó, bienven, }\\
{\color{csold}\rule[-2.5pt]{2.2pt}{9pt}}\,\textbf{\#4\ L25/f9331}\,{\color{black!60}$-1.03$}\,{\color{csold}$\phi$\,4} & \ldots{}, Ecole Normale Supérieure, Paris, France\colorbox{csold!70}{\strut \textbf{,}} 3\ldots{} & {\tiny tvguidetime, stratégique, nôtre, daction, Greif}\\
{\color{csshared}\rule[-2.5pt]{2.2pt}{9pt}}\,\textbf{\#5\ L1/f3872}\,{\color{black!60}$+1.02$} & \ldots{}\colorbox{csshared!38}{\strut Mifflin} \colorbox{csshared!33}{\strut Harcourt} \colorbox{csshared!39}{\strut Publishing} \colorbox{csshared!44}{\strut Company}\colorbox{csshared!17}{\strut .} \colorbox{csshared!40}{\strut All} \colorbox{csshared!70}{\strut \textbf{rights}} \colorbox{csshared!59}{\strut reserved}\ldots{} & {\tiny Etr, Aene, Cæsar, Eocene}\\
{\color{csold}\rule[-2.5pt]{2.2pt}{9pt}}\,\textbf{\#6\ L18/f14136}\,{\color{black!60}$-0.92$}\,{\color{csold}$\phi$\,3} & \ldots{}world (British Museum at London, Grand \colorbox{csold!48}{\strut Palais} \colorbox{csold!70}{\strut \textbf{at}} \colorbox{csold!8}{\strut Paris},\ldots{} & {\tiny WithFormat, addCriterion, uska, ainfi, excellents}\\
{\color{csshared}\rule[-2.5pt]{2.2pt}{9pt}}\,\textbf{\#7\ L0/f7081}\,{\color{black!60}$+0.84$} & <bos>\colorbox{csshared!44}{\strut Ce}\colorbox{csshared!41}{\strut va}\colorbox{csshared!70}{\strut \textbf{'}}\colorbox{csshared!53}{\strut s}\ldots{} & {\tiny EndGlobalSection, IUrlHelper}\\
{\color{csold}\rule[-2.5pt]{2.2pt}{9pt}}\,\textbf{\#8\ L19/f472}\,{\color{black!60}$+0.79$}\,{\color{csold}$\phi$\,5}\,{\color{black!60}@in} & \ldots{}Liberated and enfranchised \colorbox{csold!53}{\strut by} \colorbox{csold!49}{\strut the} events \colorbox{csold!52}{\strut of} \colorbox{csold!14}{\strut 1}\colorbox{csold!70}{\strut \textbf{7}}\colorbox{csold!55}{\strut 8}\colorbox{csold!8}{\strut 9},\ldots{} & {\tiny departments, department, Giver, Als, Seine}\\
\bottomrule
\end{tabular}
}\\[3pt]
  {\scriptsize\setlength{\tabcolsep}{3pt}
\definecolor{csshared}{HTML}{8A8985}
\setlength{\fboxsep}{0.6pt}%
\begin{tabular}{@{}l@{\hspace{5pt}}>{\raggedright\arraybackslash}p{6.75cm}>{\raggedright\arraybackslash}p{3.1cm}@{}}
\multicolumn{3}{@{}l}{\textbf{Top readouts onto ``Rome'' in $B$ (edited model)}}\\[1pt]
\toprule
{\color{csshared}\rule[-2.5pt]{2.2pt}{9pt}}\,\textbf{\#1\ L19/f9768}\,{\color{black!60}$-1.39$} & University of Tokyo\colorbox{csshared!8}{\strut \P{}}7-3-1 H\colorbox{csshared!9}{\strut ongo}\colorbox{csshared!26}{\strut ,} Bunkyo-\colorbox{csshared!46}{\strut ku}\colorbox{csshared!70}{\strut \textbf{,}} Tokyo 1\ldots{} & {\tiny úgó, bienven, }\\
{\color{csshared}\rule[-2.5pt]{2.2pt}{9pt}}\,\textbf{\#2\ L17/f11572}\,{\color{black!60}$+1.30$} & \ldots{}participating from London \colorbox{csshared!55}{\strut to} Sao Paolo\colorbox{csshared!46}{\strut ,} Seoul \colorbox{csshared!70}{\strut \textbf{to}} \colorbox{csshared!8}{\strut Babylon}\ldots{} & {\tiny Lisbon, Istanbul, Bangkok, London, Sydney}\\
{\color{csnew}\rule[-2.5pt]{2.2pt}{9pt}}\,\textbf{\#3\ L20/f4888}\,{\color{black!60}$+1.23$}\,{\color{csnew}$\phi$\,1} & \ldots{}at the Regina Elena \colorbox{csnew!8}{\strut National} Cancer Institute \colorbox{csnew!70}{\strut \textbf{in}} \colorbox{csnew!8}{\strut Rome}\colorbox{csnew!14}{\strut ,}\ldots{} & {\tiny HomeAsUp, Lati, +\#+\#, MENAFN, snippetHide}\\
{\color{csshared}\rule[-2.5pt]{2.2pt}{9pt}}\,\textbf{\#4\ L24/f8520}\,{\color{black!60}$+1.16$} & \ldots{}\colorbox{csshared!8}{\strut ratio} of the moments \colorbox{csshared!10}{\strut of} proton \colorbox{csshared!21}{\strut and} deuteron \colorbox{csshared!70}{\strut \textbf{by}} \colorbox{csshared!8}{\strut Bloch}\colorbox{csshared!50}{\strut ,}\ldots{} & {\tiny Jefus, Ans, several, myself, itself}\\
{\color{csshared}\rule[-2.5pt]{2.2pt}{9pt}}\,\textbf{\#5\ L0/f9908}\,{\color{black!60}$+1.12$} & \ldots{}makes a person feel sad and hopeless much \colorbox{csshared!70}{\strut \textbf{of}} the\ldots{} & {\tiny myself, Jefus, small, purpose, several}\\
{\color{csshared}\rule[-2.5pt]{2.2pt}{9pt}}\,\textbf{\#6\ L1/f3872}\,{\color{black!60}$+1.10$} & \ldots{}\colorbox{csshared!39}{\strut Mifflin} \colorbox{csshared!34}{\strut Harcourt} \colorbox{csshared!40}{\strut Publishing} \colorbox{csshared!45}{\strut Company}\colorbox{csshared!16}{\strut .} \colorbox{csshared!38}{\strut All} \colorbox{csshared!70}{\strut \textbf{rights}} \colorbox{csshared!58}{\strut reserved}\ldots{} & {\tiny Etr, Aene, Cæsar, Eocene}\\
{\color{csnew}\rule[-2.5pt]{2.2pt}{9pt}}\,\textbf{\#7\ L19/f4224}\,{\color{black!60}$+0.96$}\,{\color{csnew}$\phi$\,3} & \ldots{}\colorbox{csnew!44}{\strut Tra}\colorbox{csnew!24}{\strut jan}\colorbox{csnew!8}{\strut )} \colorbox{csnew!47}{\strut and} \colorbox{csnew!43}{\strut Marcia} \colorbox{csnew!32}{\strut Furn}\colorbox{csnew!36}{\strut illa} \colorbox{csnew!16}{\strut (}\colorbox{csnew!29}{\strut second} \colorbox{csnew!8}{\strut wife} \colorbox{csnew!70}{\strut \textbf{of}} \colorbox{csnew!60}{\strut the}\ldots{} & {\tiny Caesar, Fla, legion, Marcus, Roman}\\
{\color{csshared}\rule[-2.5pt]{2.2pt}{9pt}}\,\textbf{\#8\ L0/f8961}\,{\color{black!60}$+0.93$} & \ldots{}for HIV/AIDS patients. January\P{}Institute \colorbox{csshared!70}{\strut \textbf{of}} Medicine\ldots{} & {\tiny Efq, Majefty, Jefus, Monfieur, purpose}\\
\bottomrule
\end{tabular}
}
  \caption{\textbf{What the readout ranking sees, and what the diff flags.}
  (a,\,b) the two graphs of the running example. (c,\,d) the features with the largest
  direct weight $|w|$ onto the answer logit in each graph (sign shown; negative =
  suppresses), the ranking a circuit-tracer session gives with no diff. Each carries the
  dashboard's highest-activating context (shaded by activation, peak in bold) and its top
  promoted logits. Features sit at the final token unless marked \emph{@token}. Bar
  and shading carry the diff's verdict: \textcolor{csold}{blue} = disappeared,
  \textcolor{csnew}{orange} = appeared, gray = in both; $\phi\,n$ is the rank under
  $\phi^{(a)}$ on that side.}
  \label{fig:cs_readouts}
\end{figure}

Figure~\ref{fig:cs_readouts} lets a reader answer the question the method exists for,
unaided: do the nodes circuit-diff flags \emph{look like} the old and new objects, with
no label written by us or by a language model? Every feature appears exactly as the CLT's
released dashboard \cite{clt_gemma2_2b_426k} presents it (the tokens it promotes and its
highest-activating context), with the top-15 per side in
Appendix~\ref{apx:cards_eiffel}.

\subsection{Paris $\to$ Rome}\label{sec:cs_pr}
The edit landed ($M_e$'s argmax on $p$ is Rome), though the frozen lens keeps a
replacement score of only $0.49$ on $M_e$, at the low end of the post-edit range in
Figure~\ref{fig:single_edit_A_vs_B}. The two graphs differ where the edit predicts they
should. Disappeared nodes concentrate at the final position in layers 15--25
, the readout of the old answer, and at the relation tokens ``in the'';
appeared nodes form a band over layers 7--10 at the subject's last token,
exactly where MEMIT writes, plus a new late-layer column at the final position.

The top movers are recognizable without a label (Figure~\ref{fig:cs_readouts}c,\,d;
Appendix~\ref{apx:cards_eiffel}). The rank-1
disappeared node L18/f7269 promotes French tokens and fires on ``Grand Palais
\emph{at} Paris''; L20/f9908 fires on ``Louvre Museum \emph{in} Paris''. The
rank-1 appeared node L20/f4888 fires on ``\emph{in} Rome, Italy''-style
contexts, and L17/f13216 (rank 4, sitting at the subject token) promotes
``Italy'': the subject representation itself now carries the new object. The
same reading holds on the other edits we inspected: two of them in
\S\ref{sec:cs_two} below, and, for one more, the French\,$\to$\,English edit
loses ``françaises / franceses'' and ``Laurent / Sarkozy'' features.

\paragraph{Beyond the token.} On this edit the flagged features are not only detectors of
the answer string; several of them read as the surrounding idea of the place. Deeper in the appeared ranking sit a Roman-antiquity feature promoting
``Caesar / legion / Roman'' (L19/f4224, rank~3), a Roman-mythology feature firing on
``Romulus and Remus in 753'' (L20/f7760, rank~5) and a papacy feature firing on ``rival
popes \emph{in} Rome and Avignon'' (L22/f4296, rank~13); the disappeared side mirrors it
with a French Revolution feature (``the events of 1789''; L19/f472, rank~5) and French
port cities (``Havre / Lyons / Nice''; L19/f12469, rank~14). Four more of each are in
Appendix~\ref{apx:cards_eiffel}. What the diff surfaces on this edit is therefore Roman
history, mythology and the Church on one side and French history and geography on the
other, rather than the two object tokens alone.

\paragraph{Recurrence separates knowledge from mechanism.} Some top features recur across
edits; the dagger counts in how many of the 25 (the running example plus the 24 of the
sweep) a feature appears. L18/f7269$^{\dagger 4}$ recurs in exactly the other French facts
of the sweep, recurrence that tracks shared knowledge. By contrast
L7/f12622$^{\dagger 12}$, L5/f8658$^{\dagger 5}$ and L25/f9787$^{\dagger 11}$ recur across a
dozen unrelated edits with uninterpretable dashboards, and the first two sit at the
subject position inside the MEMIT band, which suggests they track the write itself rather
than the fact. A reader should discount them, and the platform reports a cross-edit
recurrence count as a cheap filter for exactly that (Appendix~\ref{apx:cards_eiffel}).

\subsection{Two further edits}\label{sec:cs_two}
We apply the same reading to two further edits from the sweep of
\S\ref{sec:exper_verif}: \emph{Taj Mahal, follower of, Islam $\to$ Christianity} and
\emph{Pentium II, developed by, Intel $\to$ Microsoft}. Both repeat the geometry of
\S\ref{sec:cs_pr} (appeared
nodes banded across the MEMIT layers at the subject's last token, disappeared nodes
columned at the final position), and both name their objects without a label: Taj Mahal
trades a ``Shah / Allah'' feature for ``Protestant / Christian / churches'', and
Pentium~II trades a ``Hyper Threading \emph{enabled} CPU'' feature for one firing inside
Transact-SQL documentation. Both diffs and their top-6 features per side are in
Appendix~\ref{apx:cs_two}.

\section{Limitations}\label{sec:limitations}

\textbf{Dependence on the edit and on the lens.} The method rests on two things: that the
factual edit succeeded, and that the base CLT remains a usable lens afterward. The weaker
the edit, the less we expect the diff to mean: if the model barely changed, few
features move, and the ones that do need not be the ones carrying the fact. We do not measure this dependence directly.

The lens is the more fundamental limitation. Holding $T_0$ fixed preserves the meaning
its features had for the base model, which is what makes the released dashboards
readable, but does not guarantee that meaning still describes $M_e$.
\S\ref{sec:cltReliablityafteredit} shows high pre/post agreement away from the fact, and
the replacement--completeness gap is consistent with error nodes absorbing what the lens
can no longer model; but that is an inference from two aggregate scores, and it bounds
how much of the knowledge-carrying circuit the method can recover.

\textbf{Practicality and scale of the evidence.} Figure~\ref{fig:setdiff_combined_null}
shows that the raw appeared/disappeared reading is too permissive to be useful on its own:
it covers most of the graph on the gained side. The null floor of Step~4 is what makes the output a short list ($14.0\%$ and $3.2\%$ of
each side), so the method's practical value rests on a calibration whose percentile we
have not tuned systematically. Single-node patches move the logit little, as
\citet{persic2025circuittracing} also report, so the causal evidence rests on group
patches over 24 edits (12 for the six-score comparison), which is enough for a per-edit
sign test but not for a tight effect estimate. And the selected sets overlap heavily with the
graph's top-influence set, so on the suppress side the advantage over simply reading off
influence is small or absent.

\textbf{Limitations inherited from the edit method.} Causal tracing and ROME/MEMIT carry
known problems that the diff inherits. Causal tracing does not reliably identify the best
MLP layer at which to overwrite a fact \citep{hase_does_2023}; longer-form evaluation of
edited models finds the edit applied inconsistently \citep{rosati_long-form_2024}; and
knowledge editing produces unwanted side effects at scale
\citep{hoelscher-obermaier_detecting_2023,yang2024understandingcollapsellmsmodel}. Any
such side effect appears in the diff as movement that is not the fact, which is one
reason the recurrence filter of \S\ref{sec:cs} is worth reporting alongside the ranking.

\textbf{Limitations of the CLT basis itself.} Cross-Layer Transcoders are now a common
tool for approximating circuits in language models, but their mechanistic faithfulness has
been questioned on toy models where ground truth is available
\citep{lange_rgrgrg_dearstyne_maher_2026}. Circuit-Diff inherits whatever the basis gets
wrong: it can only report movement among features the dictionary happens to represent.

\section{Conclusion}
We set out to make the feature nodes that carry a fact easier to find, and the simplest version of the idea already carries signal: edit the fact, rebuild the attribution graph through the same frozen CLT, and look at what moved. On the 24 edits we patch, we find that the features our method elicits present, checked from their own dashboards and ablations, as the knowledge around the object rather than just its name. We release the platform and every table behind these figures so others can try the method on their own edits and CLTs.

\begin{ack}
We thank the Carnegie Mellon University Undergraduate Research Development Center for
funding Edward G. Friedman's research.
\end{ack}


\newpage
\bibliographystyle{plainnat}
\bibliography{references}


\newpage
\appendix
\addtocontents{toc}{\protect\setcounter{tocdepth}{2}}
\tableofcontents
\newpage

\section{Defaults Used in Experiments}\label{apx:defaults}
Unless a deviation is stated where an experiment is described, all results use
the defaults in Table~\ref{tab:defaults}.
\begin{table}[h]
  \caption{Defaults used across experiments.}
  \label{tab:defaults}
  \centering
  \small
  \begin{tabular}{@{}lll@{}}
    \toprule
    Parameter & Default & Role \\
    \midrule
    \multicolumn{3}{@{}l}{\emph{Scores (\S\ref{sec:meth_Alg} and Table~\ref{tab:scores_extra})}} \\
    $\alpha$ & $1.0$ & exponent in $\phi^{(a)}, \phi^{(b)}$ (sweep: $\{0.5, 1, 2\}$) \\
    $\tau_a$ & $0$ & activation guard off; $\tfrac{0}{0}\equiv 0$ in $\delta_n$ \\
    MAD floor & $10^{-12}$ & denominator floor in $\phi^{(c)}$'s robust $z$ \\
    \midrule
    \multicolumn{3}{@{}l}{\emph{Feature-patching verification}} \\
    $\varepsilon_R$ & $0.01$ & guard for $R_{\mathrm{old}}, R_{\mathrm{new}}$ \\
    $\lambda$ grid & $\{0.5, 1, 2, 4, 6, 8\}$, $-\{0, 1, 2, 4, 6, 8\}$ & appeared, disappeared \\
    $\lambda^{*}$ & $\arg\max R^{\mathrm{adj}}_{\mathrm{new}}$ / $\arg\min R^{\mathrm{adj}}_{\mathrm{old}}$ & pre-registered per node \\
    $q$ grid & $\{0.1, 0.5, 1, 5, 10\}\%$ & selection quantiles \\
    KL ceiling & $1.0$ nat & rows above excluded \\
    \midrule
    \multicolumn{3}{@{}l}{\emph{Editing and CLT}} \\
    $\mathcal{B}$ & $[3, 10]$ & MEMIT edit band \\
    $k$ & $1$ & facts per model \\
    $\eta$ & $10^{-4}$ & CLT fine-tune LR \\
    \bottomrule
  \end{tabular}
\end{table}

\section{Compute Resources} \label{apx:compute}
Every experiment in this paper runs on a single GPU. Attribution-graph computation and
patching use one NVIDIA RTX A6000 (48\,GB) on an internal SLURM cluster, with
${\sim}96$\,GB of host memory used to stage the 426k-feature CLT before it is moved
to the device; the 48\,GB budget is the binding constraint, and is what forces the
decoder-slice paging described in Appendix~\ref{apx:clt_finetuneing}. MEMIT edits and
prompt resolution are comparatively cheap and run in the same allocation. No experiment
here requires multi-GPU or multi-node execution, and no model is trained from scratch:
the only training is the additive-delta CLT fine-tune of
Appendix~\ref{apx:clt_finetuneing}, which at $k = 1$ stops after 100 optimizer steps
($181{,}454$ tokens).
The dominant cost is the patching sweep. The 24-edit group ablation of
\S\ref{sec:exper_verif} measured $154{,}345$ patch cells for a total of
$10.3$ GPU-hours, a median of $0.42$ GPU-hours per edit (${\sim}6{,}300$ cells each), each
edit fitting in a single shard.
The full project consumed substantially more compute than the runs reported here, through
preliminary sweeps, discarded floor percentiles and score variants, and the multi-edit
ladder of Appendix~\ref{apx:clt-faithfullness_multiedit}.

\section{Released Code and Data} \label{apx:release}
The Circuit-Diff library is released as a pip-installable package
(\texttt{pip install circuit-diff}, import \texttt{circuit\_diff}, CLI \texttt{cs}) at
\url{\cdurlcode}, under the MIT license. It contains the diff computation, the node
scores of \S\ref{sec:meth_Alg}, the patching stages used in \S\ref{sec:exper_verif}, the
multi-prompt aggregation of Appendix~\ref{apx:mpd}, and a local viewer. The viewer
vendors a patched copy of Anthropic's attribution-graphs frontend (MIT, \textcopyright{}
2025 Anthropic), with the patches documented in the repository. A bundled $11$\,MB
attribution bundle for the Eiffel Paris\,$\to$\,Rome case study ships with the package,
so \texttt{cs load <bundle> --serve} reproduces Figure~\ref{fig:cs_readouts}(a,\,b) with
no computation. The public adapter targets Gemma-2-2B only. A hosted demo of the viewer runs at
\url{\cdurldemo}. A second branch,
\texttt{experiments}, adds the campaign code, run configurations and figure scripts
behind the quantitative figures (\texttt{setdiff-500}, \texttt{cf-sweep-24},
\texttt{ablation-redo}, \texttt{ablation-verification}, \texttt{score-ablation}).

The summary tables behind those figures are released separately as a dataset
($23$\,MB, $83$ files) at \url{\cdurldata} under CC-BY-4.0: the full Eiffel
attribution bundle, the per-edit set-difference and null-floor tables behind
Figure~\ref{fig:setdiff_combined_null}, the 24-edit CounterFact sweep, the group and
single-node patching tables behind Figures~\ref{fig:ablate_group_k},
\ref{fig:ablate_single} and \ref{fig:ablate_single_lambda_pair} including the
null-floor and duplicate-gate robustness variants, the rank-causality dose--response
tables, and the six-score comparison behind Figure~\ref{fig:ablate_scores_k}.

\paragraph{What is not released.} No model weights and no edited checkpoints. The MEMIT
weight deltas for the case-study edits are not part of this drop. Base weights come from
Google under the Gemma Terms of Use.

\paragraph{Reproducibility boundary.} Loading, viewing and ranking a released bundle, and
regenerating the paper's diff, patching and case-study tables from the released data,
require none of the editing stack: the library's load path depends only on
\texttt{pandas} and \texttt{pyarrow}. Computing \emph{new} bundles or edits does require a
GPU, gated access to \texttt{google/gemma-2-2b}, circuit-tracer 0.4.1 (decoderesearch
fork) and MEMIT wiring that is not part of this release. The MEMIT-reproduction figures
(Appendix~\ref{apx:memitongemma}) and the CLT lens-grid figures
(Appendices~\ref{apx:clt_finetuneing} and \ref{apx:clt-faithfullness_multiedit}) come
from a separate experimental codebase and are not regenerated by the released repository.

\section{Assets and Licenses} \label{apx:assets}
All artifacts used in this paper are publicly released and are used under their stated
terms. Gemma-2-2B \citep{team_gemma_2024} is used under the Gemma Terms of Use. The
CounterFact dataset and the ROME/MEMIT implementation
\citep{meng_locating_2023,meng_mass-editing_2023} are used under the MIT license.
circuit-tracer \citep{hanna-etal-2025-circuit}, on which our platform is built, is used
at version 0.4.1 (decoderesearch fork) under the MIT license. The 426k-feature cross-layer transcoder for Gemma-2-2B and its
feature dashboards \citep{clt_gemma2_2b_426k} are used under the terms stated on the
model's Hugging Face card. The probe-prompting role assignment of
Appendix~\ref{apx:autointerp} is adapted from \citet{birardi_automated_2026}, with the
deviations from their published procedure noted there. The Circuit-Diff platform is released under the MIT license and its
accompanying tables under CC-BY-4.0; the release vendors a patched copy of Anthropic's
attribution-graphs frontend (MIT, \textcopyright{} 2025 Anthropic). No model weights,
edited checkpoints or new datasets are released with it
(Appendix~\ref{apx:release}).

\section{Broader Impacts} \label{apx:impacts}
Circuit-Diff is a tool for locating the features a fact runs through. The intended use is
auditing and debugging: knowing where a model stores a claim is a prerequisite for
checking whether it stores it at all, for tracing an incorrect output to its source, and
for evaluating whether an edit did what its author intended. The same localization could
in principle make targeted modification of a model's factual behavior easier, including
modification a downstream user would not notice; we note that the localization signal here
is obtained \emph{from} an edit, so it does not lower the barrier to performing one.

The risk specific to this class of method is subtler than misuse: a localization can be
wrong and still look convincing. A ranked list of features with readable dashboards
invites the reader to believe a circuit has been identified when what has been identified
is movement in an approximation. We have tried to make that failure mode visible rather
than hide it, by reporting how much the frozen lens degrades after an edit
(\S\ref{sec:cltReliablityafteredit}), by showing that the unfloored reading covers most of
the graph (\S\ref{sec:exper_featurefound}), and by reporting the side on which a plain
influence ranking matches or beats our score (\S\ref{sec:exper_verif}). Interpretability
results of this kind should not be used to make claims about what a deployed model does or
does not know without independent verification.

\section{MEMIT Reproduction on Gemma-2-2B} \label{apx:memitongemma}
\begin{figure}[t]
    \centering
    \includegraphics[width=\textwidth]{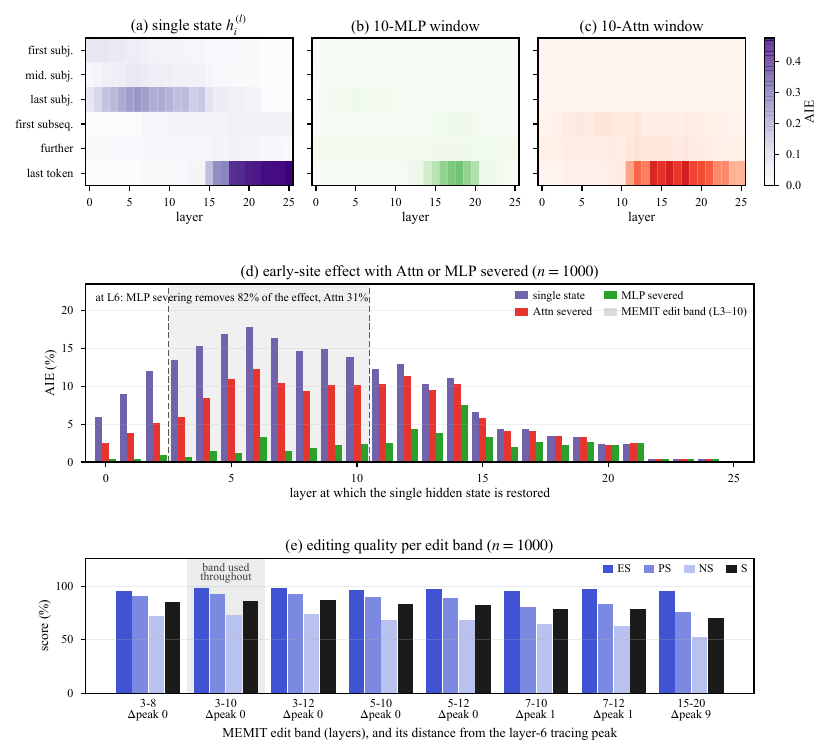}
    \caption{Causal tracing on Gemma-2-2B, and the layer-band ablation used to select the edit layers. Panels (a)--(c) reproduce the averaged causal-trace heatmaps of \citet{meng_locating_2023}, panel (d) reproduces the severed-module variant from the same work, and panel (e) reproduces Figure~11 of \citet{meng_mass-editing_2023}.
    (a) Average indirect effect (AIE) of restoring a single clean hidden state in a run where the subject tokens have been corrupted with noise.
    (b) The same measurement for restoring a window of ten consecutive MLP
    outputs, with the window center on the horizontal axis.
    (c) The same for a window of ten attention outputs. Panels (a)--(c) share a single color scale, set by panel (a), following the upstream figure.
    (d) AIE at the last subject token when either all MLP or all attention
    modules at that token are held at their corrupted values. Shading marks layers 3--10, the layer choice used for all MEMIT edits in this paper.
    (e) Editing metrics for eight candidate bands, with MEMIT applied to the
    same 1000 CounterFact records in each case, see Figure~\ref{fig:gemma-repro-editing} for the definition of each metric.}
    \label{fig:gemma-repro-tracing}
\end{figure}

We perform factual edits with ROME \citep{meng_locating_2023} and its successor MEMIT
\citep{meng_mass-editing_2023}. \citet{meng_locating_2023} introduce causal tracing, a
series of causal interventions on a model's internal activations that identifies which
layers and components are decisive for a factual prediction.
Figure~\ref{fig:gemma-repro-tracing} shows our reproduction of those causal-tracing
results on Gemma-2-2B; following their convention we edit the layer band
$\mathcal{B} = [3, 10]$, which has the largest gap between the effect of severing the MLP
and severing the attention modules. ROME and MEMIT apply low-rank updates to the
feed-forward MLPs of the layers in $\mathcal{B}$ to change the stored association. MEMIT
scales to large batches of edits, but \S\ref{sec:cltReliablityafteredit} and
Appendix~\ref{apx:clt-faithfullness_multiedit} show that the frozen transcoder degrades
sharply as the number of edits grows, so we edit one fact per model throughout.

\begin{figure}[p]
    \centering
    \includegraphics[width=\textwidth]{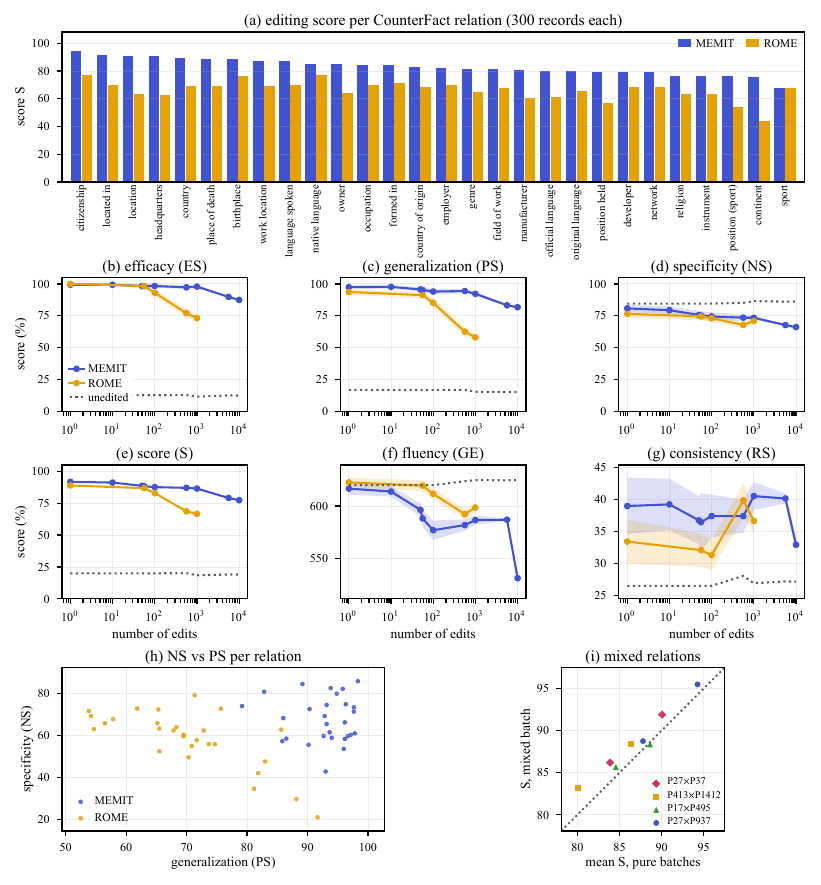}
    \caption{CounterFact editing results on Gemma-2-2B, for MEMIT applied to
    layers 3--10 and ROME applied to layer 6. Panel (a) reproduces Figure 6 of \cite{meng_mass-editing_2023}, panels (b)--(g) reproduce Figure 5, panel (h) is a second projection of the runs in panel (a), and panel (i) reproduces Figure 7.
    The metrics, from \cite{meng_locating_2023}, are as follows: ES (efficacy) is the success rate of each edit; PS (generalization) is the success rate on
    paraphrase prompts; NS (specificity) is the success rate on unrelated facts; and
    S is the harmonic mean of ES, PS and NS. 
    GE (fluency) is the weighted $n$-gram
    entropy of 100-token continuations from the generation prompts, which
    decreases as generated text becomes more repetitive. RS (consistency) is the TF-IDF cosine similarity between those continuations and reference text about the counterfactual object. 
    (a) S for each of the 28 CounterFact relations with at least 300 records. MEMIT is applied as one batch of 300 edits per relation and ROME sequentially over the same 300. Six relations with fewer than 300 records are omitted (P138, P190, P264,
    P36, P407, P463).
    (b)--(g) Each metric against the number of edits applied to the model. MEMIT
    is evaluated at $n \in \{1, 10, 50, 56, 100, 562, 1000, 5623, 10000\}$ and
    ROME at $n \in \{1, 56, 100, 562, 1000\}$. Shaded regions are 95\%
    confidence intervals over cases; dotted lines are the unedited model
    evaluated on the same records.
    (h) NS against PS for each relation, from the same runs as panel (a).
    (i) S of a batch mixing two relations against the mean S of the two
    single-relation batches, for the four relation pairs used in Appendix~D of \citet{meng_mass-editing_2023}, at two batch sizes each. The dotted line is $y=x$.
    Two deviations from the upstream protocol apply: each point in (b)--(g) is evaluated on $\max(n, 300)$ records rather than 10{,}000, and the ROME curves stop at $n{=}1000$.}
    \label{fig:gemma-repro-editing}
\end{figure}

\section{Probe-Prompting Details} \label{apx:autointerp}
All signature statistics and thresholds follow Table~3 of \citet{birardi_automated_2026}.
For feature $i$ and probe $x$, $\hat a_i(x)$ is the peak activation over non-BOS
positions and a probe is active when $\hat a_i(x) > 0$. Per feature we compute
the peak consistency $\gamma_i$ (fraction of probes containing the modal peak token in
which it is the peak), the number of distinct peak tokens, the functional-token share
$c_{\mathrm{func}}$ over active probes ($c_{\mathrm{sem}} = 1 - c_{\mathrm{func}}$),
the functional-versus-semantic margin
$\Delta_{\mathrm{fs},i} = 100\,(\hat a^{\mathrm{func}}_i - \hat a^{\mathrm{sem}}_i)/
\max(\hat a^{\mathrm{func}}_i, \hat a^{\mathrm{sem}}_i)$,
and the median sparsity $\widetilde{\mathrm{sp}}_i$ of
$\mathrm{sp}_i(x) = (\hat a_i(x) - \bar a_i(x))/\hat a_i(x)$.
Roles are assigned by whichever criterion is satisfied first, with $L$ the model depth:
\begin{equation*}
\operatorname{Role}(i) =
\begin{cases}
\text{Silent} & \text{no active probe (ours)}\\
\text{Dictionary} & \gamma_i \ge 0.80 \ \wedge\ \#\text{peaks}_i \le 1\\
\text{Say}(\cdot) & \Delta_{\mathrm{fs},i} \ge 50 \ \wedge\ c_{\mathrm{func}} \ge 0.90 \ \wedge\ \ell \ge \lceil \tfrac{7}{26}L \rceil\\
\text{Relationship} & \widetilde{\mathrm{sp}}_i < 0.45\\
\text{Concept} & \ell \le \lfloor \tfrac{3}{26}L \rfloor \ \vee\ c_{\mathrm{sem}} \ge 0.50 \ \vee\ \Delta_{\mathrm{fs},i} < 50\\
\text{Review} & \text{otherwise (excluded)}
\end{cases}
\end{equation*}
The layer cuts are their $7$ and $3$ indexed to Gemma-2-2B's $L=26$; on Gemma they
resolve to exactly those integers. Functional tokens are identified by their fixed
dictionary plus punctuation and a short-lowercase-alphabetic heuristic, and a
functional peak is resolved to its semantic target by a directional search of at most
seven tokens. Naming is role-specific: Dictionary/Concept features take their
highest-activation semantic peak token; Say features are named \texttt{Say (target)};
Relationship features are named \texttt{(token) related} over the admitted token set.
Their probe-level stability filter ($0.6$) is described in their paper but absent from
the released code that produced their tables; we default it off.

\section{Fine-Tuning the Cross-Layer Transcoder ($T_0 \to T_f$)} \label{apx:clt_finetuneing}
\begin{figure}[h]
    \centering
    \includegraphics[width=\textwidth]{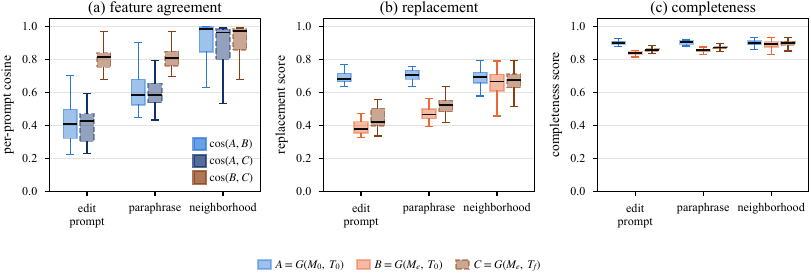}
    \caption{Fine-tuning the transcoder on the edited model does not undo the damage. Fine-tuning early-stopped at parity on all $20$ trials after $181{,}454$ tokens, so one edit does too little global damage for the objective to have anything to optimize against. The same three measures as Figure~\ref{fig:single_edit_A_vs_B}, with the adapted lens $C = G(M_e, T_f)$ added.}
    \label{fig:single_edit_finetune}
\end{figure}

\paragraph{Experiment.} We fine-tuned the CLT on $M_e$'s activations over general
Wikipedia text to obtain $T_f$. Results are in Figure~\ref{fig:single_edit_finetune}:
the replacement and completeness scores recover very little. This is largely a consequence
of how localized the edit is: there is almost nothing global for the objective to
correct. The feature agreement between $A$/$B$ and $A$/$C$ shows that the
fine-tuned lens still fires on much the same features.

We describe our training regime below: 
\paragraph{Parameterization.}
In order to train the adapted lens $T_f$, we never modify $T_0$ and instead train an additive delta over the frozen base,
$T_f = T_0 + \Delta$, where $\Delta$ contains trainable corrections only on the slices a
band-$\mathcal{B}$ edit most directly invalidates: decoder slices
$\Delta W_{\mathrm{dec}}^{(i \to l)}$ for every target layer $l \in \mathcal{B}$ and source
layer $i \le l$ (60 slices for $\mathcal{B} = [3,10]$), plus encoder and bias deltas
$\Delta W_{\mathrm{enc}}^{(l)}, \Delta b_{\mathrm{enc}}^{(l)}, \Delta b_{\mathrm{dec}}^{(l)}$
for $l \in \mathcal{B}$, giving ${\approx}2.6$\,B trainable parameters against the frozen
${426}$k-feature base. All deltas initialize to zero, so training starts from
$T_f \equiv T_0$ exactly; discarding $\Delta$ recovers the base CLT unchanged.

\paragraph{Objective and data.}
With $M_e$ as a frozen data source (its forward pass runs without gradients; autograd
touches only $\Delta$), we minimize the band-restricted CLT loss
$\mathcal{L}(\Delta) \;=\; \sum_{l \in \mathcal{B}}
  \bigl\lVert \widehat{\mathrm{mlp}}_l - \mathrm{mlp}_l \bigr\rVert_2^2
  \;+\; \beta \sum_{l \in \mathcal{B}} \lVert a^{(l)} \rVert_1$ where $\widehat{\mathrm{mlp}}_l$ is $T_f$'s reconstruction of layer $l$'s MLP output on
$M_e$, the $\mathrm{L1}$ term is per-token normalized, and $\beta = 10^{-3}$ was frozen
after a calibration benchmark chosen to keep the band's expected active features per
token within ${\sim}10\%$ of $T_0$'s. Training text is streamed from a $110$k-row Wikipedia parquet
(\texttt{wiki\_110k.parquet}) under seed $7$, with the rows of the \texttt{wiki\_v1}
evaluation battery excluded so that training and evaluation stay separate; each row
contributes BOS plus its first 128 content tokens, in batches of 16 rows with a seeded,
per-epoch-reshuffled order.

\paragraph{Optimization.}
Deltas train in bfloat16 with 8-bit paged Adam at learning rate $\eta = 10^{-4}$; decoder sources above the band are parked on CPU
between evaluations, which fits the run on a single 48\,GB GPU.

\paragraph{Stopping rule.}
Training stops at \emph{parity}: every 100 steps we evaluate $\CErec$ of $T_f$ on $M_e$
over held-out wiki rows, and stop once it reaches the value $T_0$ attains on $M_0$
(graph $A$'s faithfulness), with a hard ceiling of 30\,M tokens. This rule is
load-bearing for interpreting $C$: at $k=1$, a single edit leaves global cross-entropy
essentially intact, so \emph{every} trial ($N = 20$) passed parity at the first
evaluation: $T_f$ is $T_0$ plus exactly 100 optimizer steps ($181{,}454$ tokens).
This is a property of the stopping criterion, not of the optimizer failing to move:
at larger $k$ the parity target is not immediately reachable and training runs long
(e.g.\ ${\sim}16{,}600$ steps to the full 30\,M-token ceiling)
. We report this because Section~%
\ref{sec:cltReliablityafteredit}'s comparisons between $B$ and $C$ inherit it: at
$k=1$, $C$ is a minimally-perturbed $T_0$, which is exactly the point: the edit
gives the fine-tune almost nothing global to optimize against.

\section{CLT Faithfulness on Multi-Edit MEMIT} \label{apx:clt-faithfullness_multiedit}
\begin{figure}[p]
    \centering
    \includegraphics[width=\textwidth]{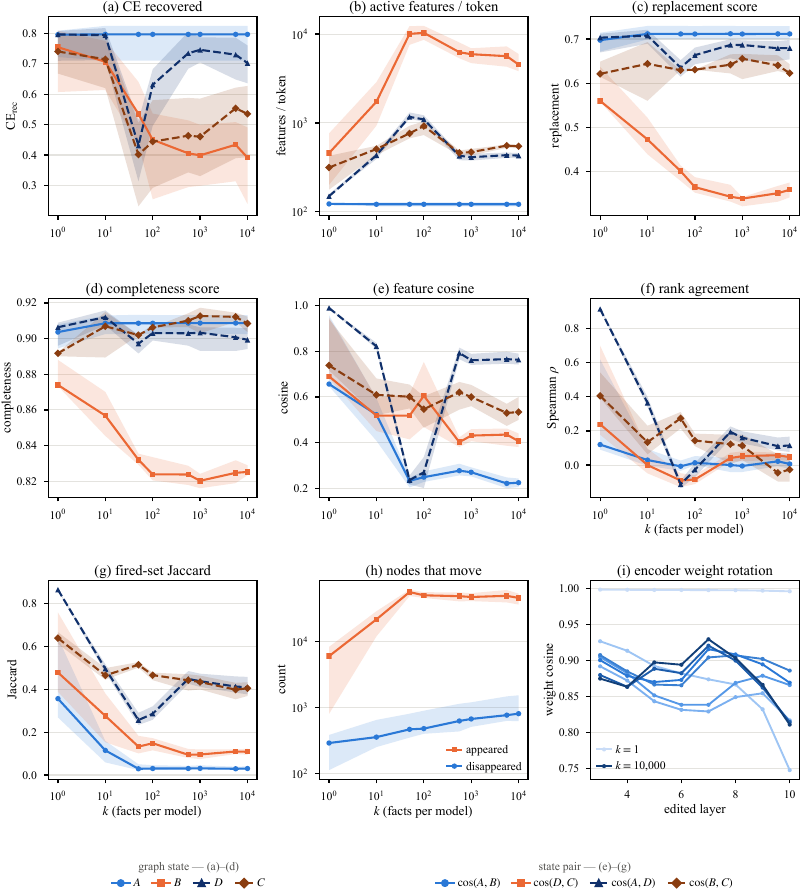}
    \caption{Multi-edit sweep across the full ladder ($k = 1$ to $10^4$ facts per model). Panels (a)--(d) report the four graph states; (e)--(g) pairwise agreement, where solid lines isolate the edit's effect (model changes, lens held) and dashed lines the lens's effect (lens changes, model held); (h) the nodes that move; (i) encoder weight rotation per edited layer, one line per $k$. Lines are medians, ribbons the interquartile range. $k = 56$ is omitted: that run carries only the $A$/$B$ arm.}
    \label{fig:multiedit_sweep_summary}
\end{figure}

Figure~\ref{fig:multiedit_sweep_summary} extends the single-edit comparison of
\S\ref{sec:cltReliablityafteredit} to $k$ facts edited into one model, over the
ladder $k \in \{1, 10, 50, 100, 562, 1000, 5623, 10^4\}$, and adds the control
corner $D = G(M_0, T_f)$: the fine-tuned lens read back on the \emph{unedited}
model, i.e.\ what fine-tuning alone does absent any edit. Three readings.
\textbf{(i)} The frozen lens degrades with $k$, and most of the damage is done by
$k=10$: on the edited prompt, $B$'s replacement score falls from $0.47$ at $k=1$ to
$0.36$ at $k=10$ and then sits near $0.35$ through $k=10^3$, while completeness moves
only from $0.85$ to $0.82$ and its active features per token rise ${\sim}6.7\times$
over the same range (panel~b): the error nodes absorb what the lens can no longer
model. \textbf{(ii)} At $k=1$, $D$ coincides with $A$
on every measure and $\cos(A, D) \approx 0.98$: fine-tuning alone changes
nothing, consistent with the parity stopping rule of
Appendix~\ref{apx:clt_finetuneing}. From $k = 50$ the fine-tune itself rewrites
the lens ($D$'s $\CErec$ drops to ${\sim}0.4$ and $\cos(A, D)$ to ${\sim}0.2$
before both recover to ${\sim}0.75$), so at large $k$ the $C$-versus-$B$
comparison is confounded by lens drift, not only by the edit; this is why the
main text holds $T_0$ fixed and uses $k=1$. \textbf{(iii)} The number of appeared
nodes saturates near $k \approx 50$ while disappeared nodes grow slowly
(panel~h), and encoder rotation concentrates in the upper edited layers
(panel~i).


\section{Additional Ablation Results} \label{apx:ablation_extra}
\begin{figure}[tp]
    \centering
    \includegraphics[width=\textwidth]{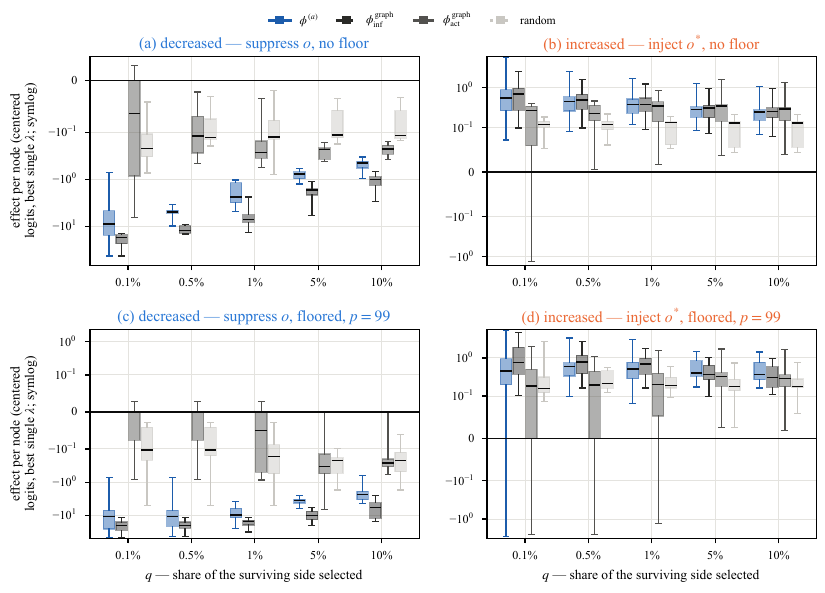}
    \caption{%
        \textbf{Patching one node at a time, under both calibrations.}
        Protocol of \S\ref{sec:exper_verif}; decreased nodes are scored against
        suppressing $o$, increased nodes against injecting $o^{*}$. Top row:
        no floor, $q$ a share of the raw side. Bottom row: only nodes that
        passed the $p=99$ null floor of Step~4, $q$ a share of the survivors.
        Arms: $\phi^{(a)}$; the whole-graph baselines
        $\phi^{\mathrm{graph}}_{\mathrm{inf}}$ and
        $\phi^{\mathrm{graph}}_{\mathrm{act}}$, which rank \emph{every} $t>0$
        node of that side's graph by influence or activation on its own target;
        and a uniform random draw. Every arm selects
        $n = \lceil q\,|\text{side}|\rceil$ nodes; each node is read at its
        best single $\lambda$ (dose--response in
        Figure~\ref{fig:ablate_single_lambda_pair}), so levels are optimistic
        bounds. Boxes are the 8 CounterFact edits, whiskers full range, symlog
        axes; the full-depth window, where circuit-tracer also freezes
        LayerNorm, is excluded for all arms.%
    }
    \label{fig:ablate_single}
\end{figure}

\begin{figure}[tp]
    \centering
    \includegraphics[width=\textwidth]{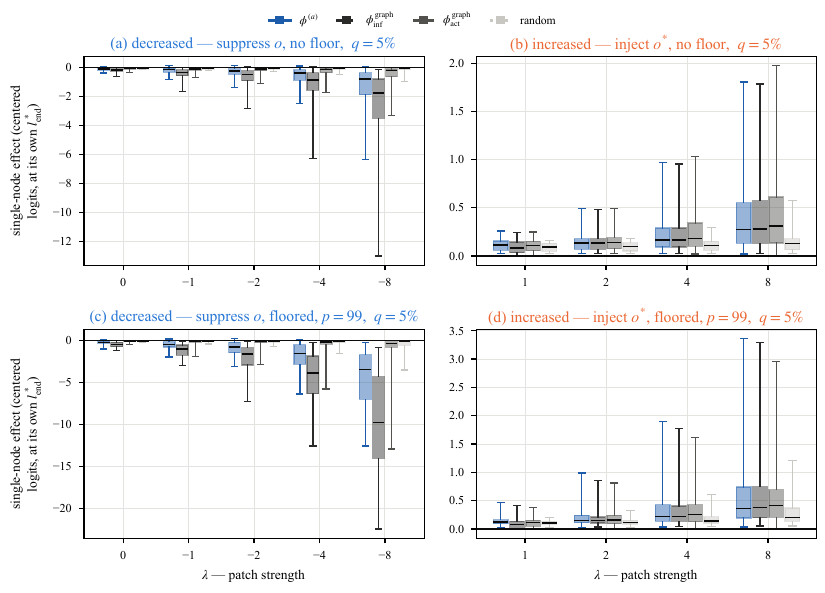}
    \caption{%
        \textbf{Dose--response of single-node patches at $q = 5\%$.}
        Same protocol, arms and edits as Figure~\ref{fig:ablate_single}; top
        row unfloored, bottom row restricted to the $p=99$ survivors. The
        grids differ by side: suppression sweeps
        $\lambda \in \{0, -1, -2, -4, -8\}$ ($0$ is the pure ablation),
        injection $\lambda \in \{1, 2, 4, 8\}$. Boxes pool nodes over 8
        CounterFact edits, whiskers 5th--95th percentile.%
    }
    \label{fig:ablate_single_lambda_pair}
\end{figure}

\begin{figure}[tp]
    \centering
    \includegraphics[width=\textwidth]{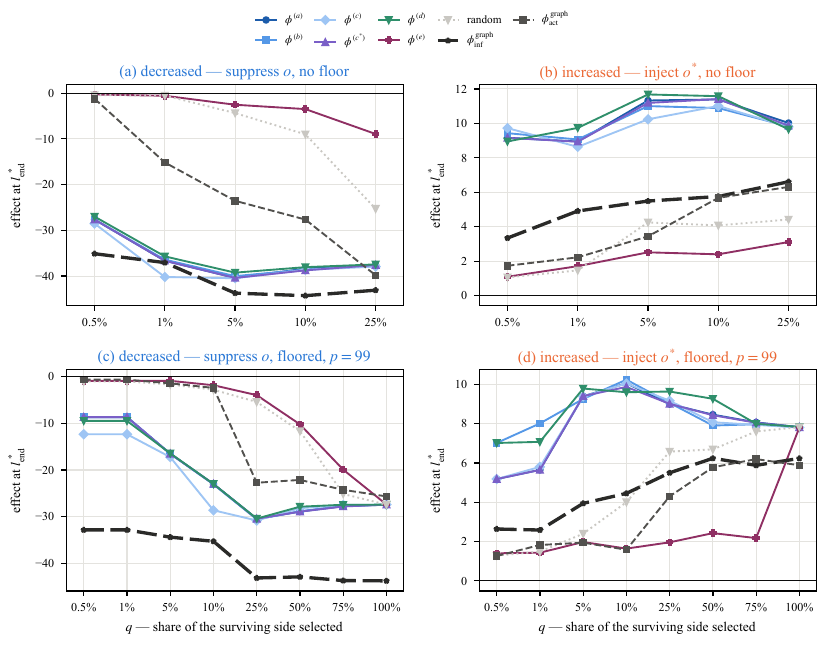}
    \caption{%
        \textbf{All six scores against the whole-graph baselines, under both
        calibrations.}
        The group ablation of Fig.~\ref{fig:ablate_group_k} on the 12 of its 24
        edits that ran every score; lines are medians over edits, member counts
        matched across all arms at every $q$. Top row: no floor. Bottom row:
        $p=99$ survivors. $\phi^{(a)}$--$\phi^{(d)}$ overplot at every rung
        under both calibrations; $\phi^{(e)}$ tracks random.%
    }
    \label{fig:ablate_scores_k}
\end{figure}

Single-node patching (Figure~\ref{fig:ablate_single}), its dose--response over the $\lambda$ grid (Figure~\ref{fig:ablate_single_lambda_pair}), and the six-score comparison (Figure~\ref{fig:ablate_scores_k}), referred to in \S\ref{sec:exper_verif}. Table~\ref{tab:scores_extra} defines the five scores compared there against $\phi^{(a)}$ of \S\ref{sec:meth_Alg}.

\begin{table}[h]
  \caption{The five further heuristic scores, ranked and patched alongside
  $\phi^{(a)}$ in Figure~\ref{fig:ablate_scores_k}. Conventions as in
  \S\ref{sec:meth_Alg}. Note that $\phi^{(b)}$, $\phi^{(c*)}$ and
  $\phi^{(d)}$ each carry $|\delta_n|$ as a magnitude, as $\phi^{(a)}$ does,
  which is why they are not separable from it: $|\delta_n|$ is saturated at
  $1$ for nearly every node on both sides.}
  \label{tab:scores_extra}
  \centering
  \small
  \begin{tabular}{@{}lll@{}}
    \toprule
    Score & Name & Definition \\
    \midrule
    $\phi^{(b)}_n$ & Weighted influence &
      $|\delta_n| \cdot \max\bigl(|I_n^{(A)}|, |I_n^{(B)}|\bigr)^{\alpha}$ \\
    $\phi^{(c)}_n$ & Influence $z$-score &
      $\bigl(|\Delta I_n| - \operatorname{Med}(|\Delta I|)\bigr) / \operatorname{MAD}(|\Delta I|)$ \\
    $\phi^{(c*)}_n$ & Strict influence $z$-score &
      $|\delta_n| \cdot \phi^{(c)}_n$ \\
    $\phi^{(d)}_n$ & Influence-weighted edge movement &
      $|\delta_n| \sum_{n'} |\Delta w_{n\to n'}| \max\bigl(|I_{n'}^{(A)}|, |I_{n'}^{(B)}|\bigr)$ \\
    $\phi^{(e)}_n$ & Edge redistribution &
      $\operatorname{JSD}\bigl(\pi_n^{(A)} \,\|\, \pi_n^{(B)}\bigr)$ \\
    \bottomrule
  \end{tabular}
\end{table}

\section{Circuit-Diff Platform} \label{apx:circuitdiffplatform}
\subsection{Multi-Prompt Aggregation}\label{apx:mpd}
To reduce the noise of relying on a single prompt, we hold the fact fixed and vary its
phrasing over $m$ prompts. Because token positions differ across phrasings, we track
features rather than nodes. Within each prompt $p$ we collapse a feature's nodes into a
single score $\Phi_i(p) = \sum_{n' \in T_i} \phi^{(s)}_{n'}$, where $T_i$ is the set of
nodes carrying feature $i$ at any token position in $p$ and $s$ is the chosen score,
normalize by that prompt's median, and take the feature's direction from the sign of its
summed $\delta_i$. We discard features whose movement is inconsistent across prompts,
i.e.\ features with both positive and negative $\delta$. We then pool across prompts with
a power mean, $\mathrm{MPD}_i = \big(\tfrac{1}{m}\sum_p \Phi_i(p)^{\rho}\big)^{1/\rho}$.
We report $\rho = \tfrac{1}{2}$ as \textsc{breadth} (rewarding features that recur across
phrasings) and $\rho = 2$ as \textsc{peak} (rewarding features that move strongly on any
one phrasing).

\subsection{Feature evidence for the case study: Eiffel Paris $\to$ Rome}\label{apx:cards_eiffel}
The listing below is the evidence behind \S\ref{sec:cs_pr}: the top-15 features
per side under $\phi^{(a)}$ for the running example (each feature at its
strongest node), with the tokens the feature promotes and its
highest-activating dashboard context (peak token in brackets), read from the
CLT's released feature dashboards \cite{clt_gemma2_2b_426k}. Nothing is
labelled by hand or by a language model. Rows are in rank order within each
side; \S\ref{sec:cs_pr} (\emph{Beyond the token}) reads the Roman-history,
mythology and Church features on the appeared side and the French-history
features on the disappeared side off this listing.

Dagger~$n$ marks a feature that is in the per-feature top-15 of either side in
$n \ge 3$ of the 25 pooled edits (the running example plus the 24 CounterFact
edits of the sweep in \S\ref{sec:exper_verif}. The count is therefore over the
whole pool, so a feature may carry a dagger without every one of its other
appearances being printed here; logit tokens in non-Latin scripts are skipped
(count in the generated file's header). Generated by \texttt{cs print-cards}
from the same \texttt{scores.py} ranking the viewer uses; the same command
prints the per-edit listing for any subset of the pool.
{\scriptsize\setlength{\tabcolsep}{4pt}
\begin{longtable}{@{}l l >{\raggedright\arraybackslash}p{3.5cm} >{\raggedright\arraybackslash}p{6.0cm}@{}}
\toprule
Side (based on $A$) & Feature & Top logits & Context \\
\midrule
\endhead
\multicolumn{4}{@{}l}{\textit{Eiffel Paris $\to$ Rome}} \\
\textcolor{csold}{disappeared} & L18/f7269$^{\dagger4}$ & \texttt{ le}, \texttt{ É}, \texttt{ Ré} &  Museum at London, Grand Palais\textbf{[ at]} Paris, National \\
\textcolor{csold}{disappeared} & L20/f9908 & \texttt{ normaux}, \texttt{TagMode}, \texttt{ financières} &  be seen at The Louvre Museum\textbf{[ in]} Paris, France \\
\textcolor{csold}{disappeared} & L18/f14136$^{\dagger3}$ & \texttt{WithFormat}, \texttt{addCriterion}, \texttt{ uska} &  of Brassempouy\textbf{[ in]} 18 \\
\textcolor{csold}{disappeared} & L25/f9331 & \texttt{tvguidetime}, \texttt{ stratégique}, \texttt{ nôtre} &  Supérieure, Paris, France\textbf{[,]} 3 Department \\
\textcolor{csold}{disappeared} & L19/f472$^{\dagger3}$ & \texttt{ departments}, \texttt{ department}, \texttt{ Giver} &  by the events of 1\textbf{[7]}89, \\
\textcolor{csold}{disappeared} & L20/f9226 & \texttt{ Valera}, \texttt{ Butte}, \texttt{ gehör} &  Lachaise Cemeteray\textbf{[ in]} Paris.\P{} \\
\textcolor{csold}{disappeared} & L12/f6700 & \texttt{ llamado}, \texttt{ stället}, \texttt{antaranya} & on National Park on the island\textbf{[ of]} Java, Indonesia \\
\textcolor{csold}{disappeared} & L15/f8344 & \texttt{complexContent}, \texttt{ Manhattan}, \texttt{///</} &  kickoff event at the Franklin Institute\textbf{[ in]} Philadelphia today showcased \\
\textcolor{csold}{disappeared} & L16/f12025 & \texttt{ardless}, \texttt{enumi}, \texttt{RectangleBorder} &  at the ceremony, Mayor of\textbf{[ Jerusalem]} Nir Barkat \\
\textcolor{csold}{disappeared} & L21/f6014 & \texttt{ Alsace}, \texttt{ departament}, \texttt{iredo} &  had advanced as far as Alsace\textbf{[-]}Lorraine, provinces \\
\textcolor{csold}{disappeared} & L7/f803 & \texttt{ Italijani}, \texttt{ItemBackground}, \texttt{+\#+} &  the middle one of which,\textbf{[ built]} of stone, \\
\textcolor{csold}{disappeared} & L20/f9342 & \texttt{\#\#\#\#\#\#\#\#.}, \texttt{expandindo}, \texttt{ estekak} &  He was superior of the seminary\textbf{[ of]} Issy when \\
\textcolor{csold}{disappeared} & L16/f14688 & \texttt{Manbalar}, \texttt{Geplaatst}, \texttt{ AssemblyTitle} &  Gemeinden in Germany...\P{}in\textbf{[ the]} Ain\P{}Ain \\
\textcolor{csold}{disappeared} & L19/f12469$^{\dagger3}$ & \texttt{ Havre}, \texttt{ Lyons}, \texttt{Constant} & ote at Marseilles---He\textbf{[ reaches]} Bombay---Letter \\
\textcolor{csold}{disappeared} & L15/f4283 & \texttt{<eos>}, \texttt{ }, \texttt{   } &  her arrangement with Hades regarding her\textbf{[ daughter]} Persephone. \\
\textcolor{csnew}{appeared} & L20/f4888 & \texttt{HomeAsUp}, \texttt{ Lati}, \texttt{+\#+\#} &  from Ascoli Satriano\textbf{[ in]} Puglia with a \\
\textcolor{csnew}{appeared} & L7/f12622$^{\dagger12}$ & \texttt{pu}, \texttt{rap}, \texttt{inge} &  that divides the waters of French\textbf{[ Broad]} and Big Pigeon \\
\textcolor{csnew}{appeared} & L19/f4224 & \texttt{ Caesar}, \texttt{ Fla}, \texttt{ legion} &  and on the mother's\textbf{[ from]} Trajan, \\
\textcolor{csnew}{appeared} & L17/f13216 & \texttt{ Italij}, \texttt{ Italians}, \texttt{Italy} &  India, France, Greece,\textbf{[ Italy]}, and the \\
\textcolor{csnew}{appeared} & L20/f7760 & \texttt{Eternal}, \texttt{Umb}, \texttt{eternal} &  the village of Acquapendente\textbf{[ in]} Tuscany by Julian \\
\textcolor{csnew}{appeared} & L20/f781 & \texttt{ XCTest}, \texttt{multer}, \texttt{ coscienza} &  National Laboratory of the Italian Institute\textbf{[ of]} Nuclear Physics, \\
\textcolor{csnew}{appeared} & L18/f802 & \texttt{ Accesat}, \texttt{ Ple}, \texttt{yntaxException} &  So for example, all Romans\textbf{[ loved]} to see people \\
\textcolor{csnew}{appeared} & L8/f14670 & \texttt{\#+\#}, \texttt{ Roskov}, \texttt{GEBURTSDATUM} &  · Washington, DC 2\textbf{[0]}201 \\
\textcolor{csnew}{appeared} & L20/f13965 & \texttt{ Split}, \texttt{ Piran}, \texttt{ Cef} &  usual big cities and spend Halloween\textbf{[ in]} Transylvania! Con \\
\textcolor{csnew}{appeared} & L20/f626 & \texttt{R}, \texttt{first}, \texttt{Rk} &  the proclaimed `wealthiest man\textbf{[ in]} Rome', was \\
\textcolor{csnew}{appeared} & L18/f6681 & \texttt{ Greco}, \texttt{ temples}, \texttt{ pyramids} & . But it is also home\textbf{[ to]} vast archeological ruins \\
\textcolor{csnew}{appeared} & L8/f14633 & \texttt{ItemBackground}, \texttt{GraphicsUnit}, \texttt{VideoCapture} &  Roberts, a biological anthropologist with\textbf{[ Glasgow]} University's \\
\textcolor{csnew}{appeared} & L22/f4296 & \texttt{avl}, \texttt{R}, \texttt{szta·} & 17)---rival popes\textbf{[ in]} Rome and Avignon \\
\textcolor{csnew}{appeared} & L16/f16014 & \texttt{ Bologna}, \texttt{ Padua}, \texttt{ Ferrara} &  other Hebrew presses were set up\textbf{[ in]} Mantua, \\
\textcolor{csnew}{appeared} & L19/f8642 & \texttt{ skydd}, \texttt{ Maggio}, \texttt{ Repubblica} & .\P{}Leonardo Campanelli of\textbf{[ the]} University of Ferrara \\
\bottomrule
\end{longtable}
}

\subsection{Two further edits: Taj Mahal and Pentium II}\label{apx:cs_two}
\begin{figure}[h]
  \centering
  \includegraphics[width=\textwidth]{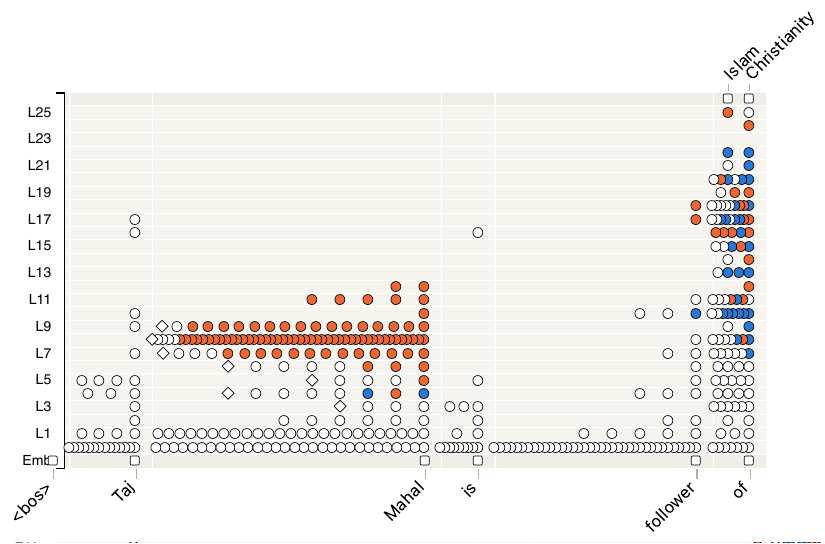}\\[-1pt]
  {\small (a) \emph{Taj Mahal is follower of}: Islam $\to$ Christianity}\\[5pt]
  \includegraphics[width=\textwidth]{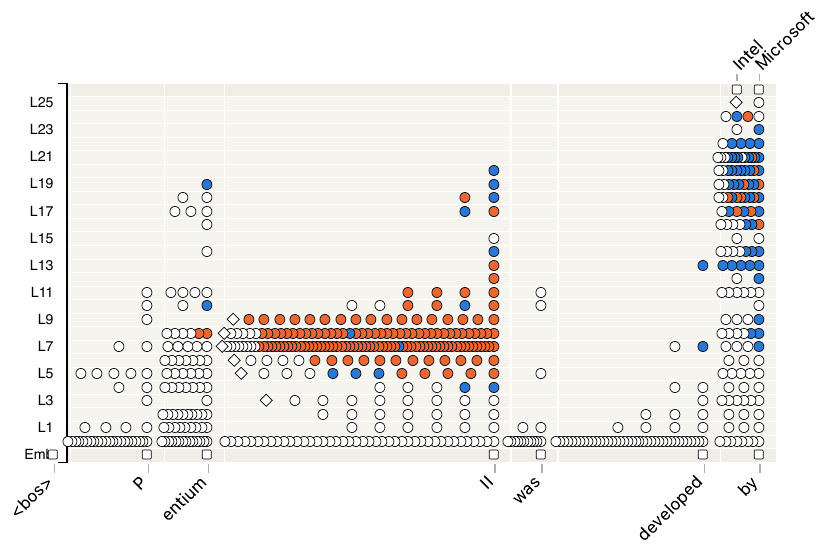}\\[-1pt]
  {\small (b) \emph{Pentium II was developed by}: Intel $\to$ Microsoft}
  \caption{\textbf{The diff on two further edits.} Conventions as in
  Figure~\ref{fig:cs_readouts}(a,\,b): columns are token positions, rows are
  layers, open markers are nodes present in both graphs, and filled markers
  carry the diff's verdict: \textcolor{csold}{blue} = disappeared,
  \textcolor{csnew}{orange} = appeared. The labelled columns at the right edge
  are the two answer logits. Both edits show the signature of
  \S\ref{sec:cs_pr}: appeared nodes form a band across the MEMIT layers at the
  subject's last token (\emph{Mahal}, \emph{II}), and the final position
  carries the readout of the old and new answers.}
  \label{fig:cs_two_graphs}
\end{figure}

The evidence behind \S\ref{sec:cs_two}, in the same form and from the same
generator as Appendix~\ref{apx:cards_eiffel}: the circuit-diff of each edit
(Figure~\ref{fig:cs_two_graphs}) and the top-6 features per side under
$\phi^{(a)}$ (listing below). Both edits are from the 24-edit sweep of
\S\ref{sec:exper_verif} (CounterFact cases 21243 and 6467); daggers are counted
over the same 25-edit pool.

{\scriptsize\setlength{\tabcolsep}{4pt}
\begin{longtable}{@{}l l >{\raggedright\arraybackslash}p{3.5cm} >{\raggedright\arraybackslash}p{6.0cm}@{}}
\toprule
Side (based on $A$) & Feature & Top logits & Context \\
\midrule
\endhead
\multicolumn{4}{@{}l}{\textit{Taj Mahal Islam $\to$ Christianity}} \\
\textcolor{csold}{disappeared} & L22/f3758 & \texttt{ Shah}, \texttt{ Allah}, \texttt{ Arjuna} &  children. It is a story\textbf{[ of]} a young boy \\
\textcolor{csold}{disappeared} & L25/f9787$^{\dagger11}$ & \texttt{ auroit}, \texttt{ varandra}, \texttt{ feroit} &  flame. In the midst of\textbf{[ this]} turbulence,  \\
\textcolor{csold}{disappeared} & L20/f65 & \texttt{ Mulai}, \texttt{ Cali}, \texttt{ Mih} &  of Spanish life has a touch\textbf{[ of]} Islam. Spanish \\
\textcolor{csold}{disappeared} & L13/f5332 & \texttt{Autoritní}, \texttt{ BoxFit}, \texttt{ myself} & <bos>The Leviathan\textbf{[ of]} Parsonstown: \\
\textcolor{csold}{disappeared} & L7/f3997 & \texttt{ must}, \texttt{ Efq}, \texttt{ chrétien} & <bos>Jesus --\textbf{[ Our]} Shepherd\P{}Week \\
\textcolor{csold}{disappeared} & L20/f7377 & \texttt{Portale}, \texttt{LookAnd}, \texttt{ ostavi} & Kalam was closely allied with\textbf{[ Islamic]} jurisprudence and typically \\
\textcolor{csnew}{appeared} & L17/f11321 & \texttt{ Protestant}, \texttt{ Christian}, \texttt{ churches} &  them - tend not to be\textbf{[ regular]} church-goers \\
\textcolor{csnew}{appeared} & L18/f842 & \texttt{Roman}, \texttt{ Lutheran}, \texttt{ Roman} & -denial traditionally observed by Catholics\textbf{[ and]} some Protestant denominations \\
\textcolor{csnew}{appeared} & L21/f10906 & \texttt{ practising}, \texttt{ faith}, \texttt{ Belie} &  prior to Christianity becoming an officially\textbf{[ tolerated]} religion of the \\
\textcolor{csnew}{appeared} & L25/f12909$^{\dagger4}$ & \texttt{.\}\textasciitilde{}\textbackslash{}}, \texttt{ividual}, \texttt{hematical} & yscale and contrast ratio mean basically\textbf{[ the]} same thing. \\
\textcolor{csnew}{appeared} & L6/f11532 & \texttt{PerformLayout}, \texttt{\$\textbackslash{}}, \texttt{ \$\textbackslash{}\$} & <bos>Summary\textbf{[:]} As part of \\
\textcolor{csnew}{appeared} & L15/f5748 & \texttt{ Judaism}, \texttt{ Byzantium}, \texttt{ Arabs} &  fine arts developed in strong interaction\textbf{[ with]} European art, \\
\addlinespace
\multicolumn{4}{@{}l}{\textit{Pentium II Intel $\to$ Microsoft}} \\
\textcolor{csold}{disappeared} & L20/f13952 & \texttt{ iVar}, \texttt{ BorderRadius}, \texttt{ Centr} &  processors or a Hyper Threading\textbf{[ enabled]} CPU can be \\
\textcolor{csold}{disappeared} & L20/f8977 & \texttt{ poveznice}, \texttt{tagHelperRunner}, \texttt{ConstraintMaker} &  relative quietness, the Taiwanese\textbf{[ manufacturer]} Winbond recently \\
\textcolor{csold}{disappeared} & L17/f13795 & \texttt{ architectures}, \texttt{ architecture}, \texttt{archite} & , Intel's x8\textbf{[6]} processor architecture is \\
\textcolor{csold}{disappeared} & L18/f10518 & \texttt{stateProvider}, \texttt{IConfiguration}, \texttt{MODO} & CL, and updated versions of\textbf{[ Intel]}'s Fortran \\
\textcolor{csold}{disappeared} & L18/f909 & \texttt{ Think}, \texttt{kt·r}, \texttt{offsetof} &  to \_OS is that it\textbf{[']}s a single \\
\textcolor{csold}{disappeared} & L14/f1292 & \texttt{ SAP}, \texttt{ Intel}, \texttt{Microsoft} &  as to persuade rivals Novartis\textbf{[ and]} Sanofi to \\
\textcolor{csnew}{appeared} & L18/f4221 & \texttt{pherson}, \texttt{ Chwiliwch}, \texttt{StructEnd} &  use in the report, as\textbf{[ well]} as calculated fields \\
\textcolor{csnew}{appeared} & L5/f11228 & \texttt{iStock}, \texttt{bootstrapcdn}, \texttt{ These} &  KEY constraint using Transact-\textbf{[SQL]} or SQL- \\
\textcolor{csnew}{appeared} & L13/f6564 & \texttt{ AssemblyCulture}, \texttt{AddTagHelper}, \texttt{EndProject} &  Journal of Environmental Research and Public\textbf{[ Health]}. In this \\
\textcolor{csnew}{appeared} & L10/f12206 & \texttt{Windows}, \texttt{Microsoft}, \texttt{ Microsoft} &  on missions beyond Earth orbit.\textbf{[\P{}]}Well, this \\
\textcolor{csnew}{appeared} & L8/f14613 & \texttt{ purpose}, \texttt{ itself}, \texttt{ Jefus} &  by creating a simple VB Windows\textbf{[ form]} project, and \\
\textcolor{csnew}{appeared} & L7/f3241 & \texttt{jwt}, \texttt{ cherchés}, \texttt{ propOrder} &  from both human rights groups and\textbf{[ technolog]}ists.\P{} \\
\bottomrule
\end{longtable}
}

\end{document}